\documentclass[journal]{IEEEtran}
\usepackage{blindtext}
\usepackage{cite}
\usepackage[pdftex]{graphicx}
\DeclareGraphicsExtensions{.pdf,.jpeg,.png}
\usepackage{algorithm}
\usepackage[noend]{algpseudocode}
\usepackage[tight,footnotesize]{subfigure}
\usepackage{caption}
\usepackage{color}
\usepackage[font=footnotesize]{subfig}
\usepackage{url}
\usepackage{amssymb}
\usepackage{amsmath}
\usepackage{mathtools}
\usepackage{makecell}
\usepackage{xcolor}
\usepackage{tikz}
\usepackage{booktabs}
\usepackage{multirow}

\newcommand{\argmin}[1]{\underset{#1}{\operatorname{arg}\,\operatorname{min}}\;}

\newcolumntype{P}[1]{>{\centering\arraybackslash}p{#1}}
\newcolumntype{M}[1]{>{\centering\arraybackslash}m{#1}}
\newcolumntype{N}{@{}m{0pt}@{}}

\makeatletter
\def\BState{\State\hskip-\ALG@thistlm}
\makeatother

\begin{document}
\title{DeepCABAC: A Universal Compression Algorithm for Deep Neural Networks}

\author{Simon Wiedemann$^\dag$, Heiner Kirchhoffer$^\dag$, Stefan Matlage, Paul Haase, Arturo Marban, Talmaj Marinc,\\ David Neumann, Tung Nguyen,  Ahmed Osman, Detlev Marpe$^*$,~\IEEEmembership{Fellow,~IEEE}, Heiko Schwarz,\\ Thomas Wiegand,~\IEEEmembership{Fellow,~IEEE} and Wojciech Samek$^*$,~\IEEEmembership{Member,~IEEE}
\thanks{This research was partly supported by the German Ministry for Education through the Berlin Big Data Center under Grant 01IS14013A and the Berlin Center for Machine Learning under Grant 01IS18037I.}
\thanks{$^*$\textit{(Corresponding authors: Detlev Marpe , Wojciech Samek.)}}
\thanks{$^\dag$Equal contribution. All authors are with Fraunhofer Heinrich Hertz Institute, 10587 Berlin, Germany (e-mail: detlev.marpe@hhi.fraunhofer.de \& wojciech.samek@hhi.fraunhofer.de).}
}

\markboth{Wiedemann et al. - DeepCABAC: A universal compression algorithm for DNNs}
{Wiedemann et al. - DeepCABAC: A universal compression algorithm for DNNs}

\maketitle

\begin{abstract}
The field of video compression has developed some of the most sophisticated and efficient compression algorithms known in the literature, enabling very high compressibility for little loss of information. Whilst some of these techniques are domain specific, many of their underlying principles are universal in that they can be adapted and applied for compressing different types of data.
In this work we present DeepCABAC, a compression algorithm for deep neural networks that is based on one of the state-of-the-art video coding techniques. Concretely, it applies a Context-based Adaptive Binary Arithmetic Coder (CABAC) to the network's parameters, which was originally designed for the H.264/AVC video coding standard and became the state-of-the-art for lossless compression. Moreover, DeepCABAC employs a novel quantization scheme that minimizes the rate-distortion function while simultaneously taking the impact of quantization onto the accuracy of the network into account.
Experimental results show that DeepCABAC consistently attains higher compression rates than previously proposed coding techniques for neural network compression. For instance, it is able to compress the VGG16 ImageNet model by x63.6 with no loss of accuracy, thus being able to represent the entire network with merely 8.7MB. The source code for encoding and decoding can be found at \url{https://github.com/fraunhoferhhi/DeepCABAC}.

\end{abstract}


\IEEEpeerreviewmaketitle

%%%%%%%%%%%%%%%%%%%%%%%%%%%%%%%%%%%%%%%%%%
% PAPER
%%%%%%%%%%%%%%%%%%%%%%%%%%%%%%%%%%%%%%%%%
\section{Introduction}
It has been well established that deep neural networks excel at solving many complex machine learning tasks \cite{DeepLearning, Schmidhuber14_DL_survey}. Their relatively recent success can be attributed to three phenomena: 1) access to large amounts of data, 2) researchers having designed novel optimization algorithms and model architectures that allow to train very deep neural networks, 3) the increasing availability of compute resources \cite{DeepLearning}. In particular, the latter two allowed machine learning practitioners to equip neural networks with an ever-growing number of layers and, consequently, to consistently attain state-of-the-art results on a wide spectrum of complex machine learning tasks.

However, this has triggered an exponential growth in the number of parameters these models entail over the past years \cite{dnn_energy_memory_gap}. Trivially, this implies that the models are becoming more and more complex  in terms of memory. This can become very problematic since it does not only imply  higher memory requirements, but also slower runtimes and high energy consumption \cite{Horowitz}. In fact, IO operations can be up to three orders of magnitude more expensive than arithmetic operations. Moreover, \cite{dnn_energy_memory_gap} show that the memory-energy efficiency trends of most common hardware platforms are not able to keep up with the exponential growth of the neural networks' sizes, thus expecting them to be more and more power hungry over time.

In addition, there has also been an increasing demand on deploying deep models to resource constrained  devices such as mobile or wearable devices  \cite{mobilenet, DL_mobile_applications, DL_mobile_wireless_networks}, as well as on training deep neural networks in a distributed setting such as in federated learning \cite{federated_learning, SatArXiv18, SatArXiv19}, since these approaches have direct advantages with regards to privacy, latency and efficiency issues. High memory complexity greatly complicates the applicability of neural networks for those use cases, in particular for the federated learning case since the parameters of the networks are transmitted through communication channels with limited capacity.

Model compression is one possible paradigm to solve this problem. Namely, by attempting to maximally compress the information contained in the network's parameters we automatically leave only the bits that are necessary for solving the task. Thus, in principle, the memory complexity of deep models should only increase with the complexity of the learning task and not with its number of parameters\footnote{To be more precise, the memory complexity of the model should only increase sublinearly with its number of parameters \cite{MDL}.}. In addition, model compression has direct practical advantages such as reduced communication and compute cost \cite{DLC_survey, DLC_survey2, Simon_lossless_dnn_compression1}. In fact, the Moving Picture Expert Group (MPEG) of the International Organization of Standards (ISO) has recently issued a call on neural network compression \cite{MPEG_nncompression_call}, which stresses the relevance of the problem and the broad interest by the industry to find practical solutions.

\subsection{Entropy coding in video compression}
The topic of signal compression has been long studied and highly practical and efficient algorithms have been designed. State-of-the-art video compression schemes like H.265/HEVC \cite{SzeBudSul_2014} employ efficient entropy coding techniques that can also be used for compressing deep neural networks. Namely, the Context-based adaptive binary arithmetic coding (CABAC) engine \cite{CABAC} provides a very flexible interface for entropy coding that can be adapted to a wide range of applications. It is optimized to allow high throughput and a high compression ratio at the same time. In particular, the transform coefficient coding part of H.265/HEVC contains many interesting aspects that might be suitable for compressing deep neural network. Hence, it appears only natural to try to adapt current state-of-the-art compression techniques such as CABAC to deep neural networks and accordingly compress them.

\subsection{Contributions}
Our contributions can be summarized as follows:
\begin{enumerate}
\item We adapt CABAC for the task of neural network compression. To the best of our knowledge, we are the first in applying state-of-the-art coding techniques from video compression to deep neural networks.
\item We quantize the parameters of the networks by minimizing a generalized form of a rate-distortion function which takes the impact of quantization on the accuracy of the network into account.
\item In our experiments we show that DeepCABAC is able to attain very high compression ratios and that it consistently attains a higher compression performance than previously proposed coders.
\end{enumerate}

\subsection{Outline}
In section \ref{sec: source coding} we start by reviewing some basic concepts from information theory, in particular from source coding theory. We also highlight the main difference between the classical source coding and the model compression paradigms in subsection \ref{subsec: model compression}. Subsequently, we proceed by explaining DeepCABAC in section \ref{sec: DeepCABAC}.  In section \ref{sec: related work} we provide a comprehensive review of the related work on neural network compression.  Finally, we provide experimental results and a respective discussion in section \ref{sec: experiments}.

\section{Source coding}
\label{sec: source coding}
%\[-------------\]
%1) lossless coding: a) sources of data $=$ probability distributions $=$ information $=$ (minimum)bits. b) lossless compression $=$ entropy limit. c) Scalar Huffman code 1 bit redundant per symbol. d) Block Huffman code or Arithmetic codes reach fundamental limit $=$ less than 1 bit per symbol.
%
%2) lossy coding: a) rate-distortion quantization. b) scalar quantization ($=$ only a bit redundant, but less complex). c) lloyd algorithm $=$ rate estimated by entropy $=$ K-Means for $\lambda=0$.
%
%Points of this section:
%\begin{enumerate}
%\item Minimum is rate-distortion curve. But scalar quantization is ok.
%\item There can be less than 1 bit per symbol. Arithmetic coders achieve that in an efficient manner.
%\item Probability distributions of sources are usually unknown! Therefore universal codes. Desiderata: codes that are flexible in that they can estimate a big set of empirical probability distributions.
%\item CABAC
%\end{enumerate}
%
%\[-------------\]

\begin{figure}[t]
\centering
\includegraphics[width=0.75\columnwidth,clip,keepaspectratio]{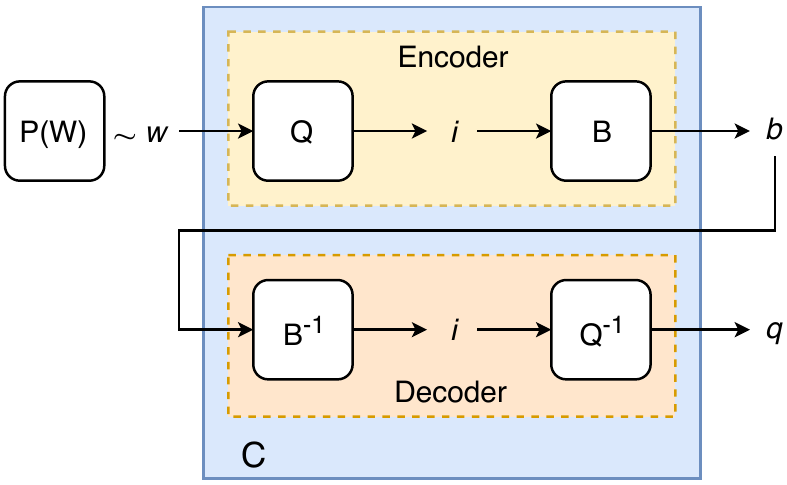}
\caption{The general structure of \textit{codes}. Firstly, the encoder maps an input sample $w$ from a probability source $P(w)$ to a binary representation $b$ by a two-step process. It quantizes the input by mapping it to an  integer $i = Q(w)$. Then, the integer is mapped to its corresponding binary representation $b = B(i)$ by applying a binarization process.  The decoder functions analogously, it maps the binary representation back to its integer value by applying the inverse $B^{-1}(b)= i$, and subsequently it assigns a reconstruction value (or quantization point) $Q^{-1}(i)= q$ to it. We stress that $Q^{-1}$ does not have to be the inverse of $Q$.}
\label{Fig: source coding}
\end{figure}

Source coding is a subfield of information theory that studies the properties of so called \textit{codes}. These are mappings that assign a binary representation and a reconstruction value to a given input element. Figure 1 depicts their most common structure. They are comprised of two parts, an \textit{encoder} and a \textit{decoder}. The encoder is a mapping that assigns a binary string of finite length $b$ to an input element $w$. In contrast, the decoder assigns a reconstruction value $q$ to the corresponding binary representation. We will also sometimes refer to $q$ as a \textit{quantization point}. Furthermore, it is assumed that the output elements $b$ and $q$  of the code $C$ are elements of finite countable sets, and that there is a one-to-one correspondence between them. Therefore, without loss of generality, we can decompose the encoder into a  \textit{quantizer} and a \textit{binarizer}, where the former maps the input to an integer value $Q(w) = i\in \mathbb{Z}$, and the latter maps the integers to their corresponding binary representation $B(i)=b$. Analogously for the decoder. Naturally, it follows that the binarizer is always a  bijective map, thus $(B^{-1} \circ B)(i) = i$.

We also distinguish between two types of codes, the so called \textit{lossless codes} and \textit{lossy codes}. They respectively correspond to the cases where $Q$ is either bijective or not, thus, the latter implies that information is lost in the coding process. Therefore, we stress that the map $Q^{-1}$ does \textbf{not} necessarily have to be the \textbf{inverse} of $Q$!

After establishing the basic definition of codes, we will now formalize  the source coding problem. In simple terms, source coding studies the problem of
\begin{quote}
\textit{finding the code that maximally compresses a set of input samples, while maintaining the error between the input and reconstruction values under an error tolerance constraint}.
\end{quote}

Notice that the problem is probabilistic in its nature since it implicitly assumes that the decoder has no access to the element values being encoded. Moreover, the input values themselves may come from an unknown source distribution. Hence, we denote with $P_{\text{Enc}}(w)$ the  encoders probability model of $w$, and with $P_{\text{Dec}}(q)\;  (\equiv P_{\text{Dec}}(b) \equiv P_{\text{Dec}}(i))$ the decoders probability model of $q$ (or equivalently $b$ and $i$). It is important to stress that both models do not have to coincide, thus $P_{\text{Enc}}(Q(w)) = P_{\text{Enc}}(i) \not\equiv  P_{\text{Dec}}(i)$. Furthermore, we will assume that the encoder's probability model follows the true underlying distribution of the input source, and therefore we will simply write $P_{\text{Enc}}(w) \equiv P(w)$.

Thus, the source coding problem can be formulated more precisely as follows: let $\mathbb{W}\subset \mathbb{R}^n$ be a given input set and let $P(w)$ be the probability of an element $w\in \mathbb{W}$ being sampled. Then, find a code $C^*$ that
\begin{equation}
C^* = \argmin{C} \mathbb{E}_{P(w)}[D(w,q) + \lambda L_C(b)]
\label{Eq: operational rate-distortion function}
\end{equation}
where $b=(B\circ Q)(w), \; q = (Q^{-1}\circ Q)(w)$. $D$ is some distance measure and $L_C$ the length of the binary representation $b$. We will sometimes refer to $L_C(\cdot)$ as the \textit{code-length} of a sample, and  with $D$ the \textit{distortion} between $w$ and $q$.  $\mathbb{E}_P[\cdot]$ denotes expectations as taken by the probability distribution $P$. $\lambda \in \mathbb{R}$ is the Lagrange multiplier that controls the trade-off between the compression strength and the error incurred by it.

Minimization objectives of the form \eqref{Eq: operational rate-distortion function} are called \textit{rate-distortion objectives} in the source coding literature. However, solving the rate-distortion objective for a given input source is most often NP-hard, since it involves finding optimal quantizers $Q$, binarizers $B$ and reconstruction values $Q^{-1}$ from the space of all possible maps. However, concrete solutions can be found for special cases, in particular in the lossless case. In the following we will review some of the fundamental theorems of source coding theory and introduce state-of-the-art coding algorithms that produce binary representations with minimal redundancy.

\subsection{Lossless coding}
\label{subsec: lossless coding}

Lossless coding implies that $q=(Q^{-1}\circ Q)(w) = w \; \forall w$. Thus, $D(w,q)=0 \;  \forall w \in \mathbb{W}$ in \eqref{Eq: operational rate-distortion function} and the rate-distortion objective simplifies into finding a binarizer $B^*$ that maximally compresses the input samples. Hence, throughout this section we will equate the general code $C$ with the binarizer $B$ and refer to it accordingly. Moreover, we will also assume that the decoder's probability model equals the encoder's, thus $P_{\text{Enc}} = P_{\text{Dec}}$. In the next subsection we discuss the case when the latter property does not apply.

Information theory already makes concrete statements regarding the minimum information contained in a probability source. Namely, Shannon in its influential work \cite{Shannon} stated that the minimum information required to fully represent a sample $w$ that has probability $P(w)$ is of $-\log_2{P(w)}$ bits. Consequently, the entropy $H_P(\mathbb{W}) = \sum_{w\in \mathbb{W}} -P(w)\log_2{P(w)}$ states the minimum average number of  bits required to represent any element $w\in \mathbb{W} \subset \mathbb{R}^n$. This implies that
\begin{equation}
H_P(\mathbb{W}) < \bar{L}_C(\mathbb{W})
\label{Eq: Fundamental theorem lossless coding}
\end{equation}
where $\bar{L}_C(\mathbb{W}) = \sum_{w\in \mathbb{W}} P(w)L_C(w)$ is the average code-length that any code $C$ assigns to each element $w\in \mathbb{W}$.  Eq. \eqref{Eq: Fundamental theorem lossless coding} is also referred as the \textit{fundamental theorem of lossless coding}.

Fortunately, from the source coding literature \cite{Wiegand_source_coding}  we know of the existence of codes that are able to reach average code-length of up to only 1 bit of redundancy to the theoretical minimum. That is,
\begin{equation}
\exists C: H_P(\mathbb{W}) < \bar{L}_C(\mathbb{W})\leq H_P(\mathbb{W}) + 1
\label{Eq: Minimal average code-length}
\end{equation}
Moreover, we even know how to build them.

Before we start discussing in more detail some of these codes we want to recall an important property of joint probability distributions. Namely, due to their sequential decomposition property, we can express the minimal information entailed in the output sample $w\in \mathbb{R}^n$ of a joint probability distribution $P(w)$ sequentially as
\[-\log_2{P(w)} = -\sum_{j=0}^{n-1} \log_2P(w_j|w_{j-1},...,w_0)\]
That is, we can always interpret a given input vector as an $n$-long random process and encode its outputs sequentially. As long as we know the respective conditional probability distributions, we can optimally encode the entire sequence. Respectively, we denote with $w_j$ the scalar value of the $j$-th dimension of $w$ (or equivalently $j$-th output of the random process). Also, we denote with $\mathbb{\mathbb{W}}_s$ the set of possible scalar inputs, where $w_j\in \mathbb{\mathbb{W}}_s, \forall j$.

\subsubsection{(scalar) Huffman coding}
One optimal code is the well-known Huffman code \cite{Huffman}. It consists of building a binary tree such that each input sample $w$ is associated with one of the leaves of the tree. Thus, each $w$ can be associated with the sequence of binary decisions that traverse the tree from its root point. The main idea is then to build the tree in such a manner that shorter paths are associated to more probable samples $w$. Huffman successfully proved that this code satisfies \eqref{Eq: Minimal average code-length}.  We provide a pseudocode of the encoding and decoding process in the appendix (see algorithms \ref{algorithm:generating-huffman-code}, \ref{algorithm:huffman-encoding}, and \ref{algorithm:huffman-decoding}).

However, Huffman codes can be very inefficient in practice since the Huffman-tree grows very quickly for large input dimensions $n$. Therefore, most often \textit{scalar Huffman codes} are used instead. Scalar Huffman codes do only consider 1-dimensional inputs, and do accordingly encode each sample from the $n$-long random process. However, these codes are suboptimal in that they produce redundant binary representations and do therefore not satisfy \eqref{Eq: Minimal average code-length}. Concretely, they produce average code-lengths of
\begin{equation*}
 H_{P}(\mathbb{W}_s) < \bar{L}_{\text{SH}}(\mathbb{W}_s)\leq H_{P}(\mathbb{W}_s) + n
\end{equation*}
where now $P(w_j)$ is the probability of a scalar output $w_j$ and $\bar{L}_{\text{SH}}(\cdot)$ is the average code-length produced by the scalar Huffman code. Moreover, they are limited to stationary processes since they do not take conditional dependencies into account, which could further reduce the average code-length.

\subsubsection{Arithmetic coding}
\begin{figure}[t]
\centering
\includegraphics[width=1\columnwidth,clip,keepaspectratio]{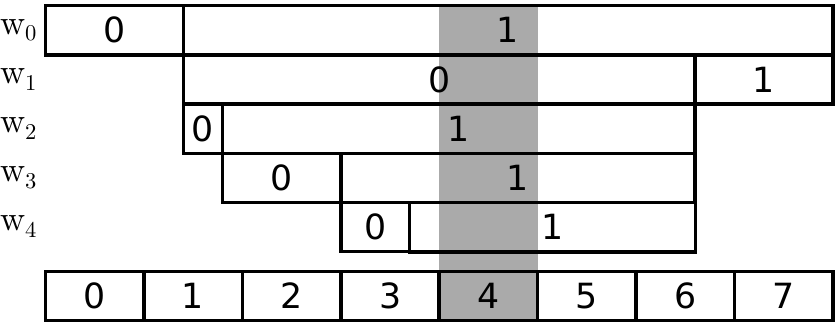}
\caption{Arithmetic coding example: Sequence '10111' is encoded as three bit sequence '100' (index '4'). The decoder can reconstruct sequence '10111' by simply selecting subintervals (for $w_0, ..., w_4$) where the coding interval is fully contained in. }
\label{Fig: arithmetic coding}
\end{figure}

A concept that approaches the joint entropy $H(\mathbb{W})$ of eq. \eqref{Eq: Minimal average code-length} in a practical and efficient manner is arithmetic coding. It consists of expressing a particular sequence of samples $w_0, w_1, ..., w_{n-1}$ of an $n$-long random process as a so called coding interval. An overview of the idea is given in the following.

Let $[L_j,L_j+R_j)$ be the coding interval before encoding symbol $w_j$ and let $L_0=0$ and $R_0=1$. Encoding of a symbol $w_j$ corresponds to deriving a coding interval $[L_{j+1},L_{j+1}+R_{j+1})$ from the previous interval $[L_j,L_j+R_j)$ as follows. Subdivide $[L_j,L_j+R_j)$ into one subinterval for each element $w_j$ of $\mathbb{W}_s$ so that the interval width is given as
\[
R_j\cdot P(w_j|w_{j-1}, w_{j-2}, ..., w_0)
\]
for a given sequence of (already sampled) values $w_{j-1}, w_{j-2}, ..., w_0$, and arrange the subintervals so that they are non-overlapping and adjacent. The subinterval associated with the sample $w_j$ to be encoded becomes the new coding interval $[L_{j+1},L_{j+1}+R_{j+1})$. Encoding of $n$ symbols yields the coding interval $[L_n,L_n+R_n)$ and the sequence of symbols $w_0, w_1, ..., w_{n-1}$ can be reconstructed (in the decoder) when an arbitrary value inside of this coding interval is known. Figure \ref{Fig: arithmetic coding} exemplifies this procedure for a binary random process. Interestingly, the width of the coding interval $R_n=P(w_0, w_1, ..., w_{n-1})$ equals the probability of sequence $w_0, w_1, ..., w_{n-1}$. As the minimum achievable code length for encoding of the $n$ symbols is known to be $-\log_2({R_n})$, the location of interval $[L_n,L_n+R_n)$ needs to be signaled to the decoder in a way so that the number of written bits gets as close to $-\log_2({R_n})$ as possible. The basic encoding principle is as follows. Derive an integer $k$ so that
\begin{equation}
2^k\le \frac{R_n}{2}<2^{1-k}
\label{Eq: arithmetic coding inequality}
\end{equation}
holds. Subdivide the unit interval $[0,1)$ into $2^k$ (adjacent and non-overlapping) subintervals $[q2^k,(q+1)2^k)$ of width $2^k$. Equation \eqref{Eq: arithmetic coding inequality} guarantees that one of the intervals $[q2^k,(q+1)2^k)$ is fully contained in the coding interval (regardless of the exact location $L_n$ of interval $[L_n,L_n+R_n)$) and if the decoder knows this interval, it can unambiguously identify $[L_n,L_n+R_n)$. Consequently, the index $q$ identifying this interval is written to the bitstream using $k$ bits.
Equation \eqref{Eq: arithmetic coding inequality} can be rewritten as
\begin{equation}
k<-\log_2({R_n})+2
\label{Eq: arithmetic coding redundancy}
\end{equation}
which shows that the ideal arithmetic coder only requires up to two bits more than the minimum possible code length for a sequence of length $n$.

\subsection{Universal coding}
\label{subsec: universal coding}

In the previous subsection we learned that there exist codes that are able to produce binary representations of (almost) minimal redundancy (e.g. arithmetic codes). However, recall that the decoder has to know the joint probability distribution of the input source in order to build the most optimal binary representation. However, in most practical situations the decoder has no prior knowledge about it. Hence, in such cases, we have to rely on so called \textit{universal codes}. They basically apply the following principle: 1) start with a general, data-independent probability model $P_{\text{Dec}}$, 2) update the model upon seeing incoming samples, 3) encode the input samples with regards to the updated probability model.

Thus, the theoretical minimum of universal codes is upper bounded by the decoder's probability estimate. Concretely, let  $P_{\text{Dec}}$ be the decoder's estimate of the input's probability model, then the minimum average code-length that can be achieved is
\[ \bar{L}_C(\mathbb{W}) > H_{P,P_{\text{Dec}}}(\mathbb{W}) = H_P(\mathbb{W}) + D_{KL}(P \, || \, P_{\text{Dec}})\]
with $H_{P,P_{\text{Dec}}}(\mathbb{W}) = -\sum_{w\in \mathbb{W}} P(w)\log_2{P_{\text{Dec}}(w)}$ being the cross-entropy and $D_{KL}$ the Kullback-Leibler divergence. Hence, a lossless code can only create binary representations with minimal redundancies iff its decoder's probability model is the same as the input sources. In other words, the better its estimate is, the better it can encode the input samples.

An example of a universal lossless code is the so called \textit{two-part Huffman code}. Given a set of samples to be encoded, it firstly makes an estimate of their empirical probability mass distribution (EPMD) and, subsequently, it encodes the samples with regards to it. However, it has the natural caveat that the estimate needs to be encoded as well, which may add a significant number of bits in many practical situations. Moreover, as we already discussed in the previous subsection, Huffman codes also come with a series of undesired properties that make it very inefficient for cases where fast adaptability and coding efficiency is required \cite{Wiegand_source_coding}.

In general, a universal lossless code should have the following desiderata:
\begin{itemize}
\item \textbf{Universality:} The code should have a mechanism that allows it to adapt its probability model to a wide range of different  types of input distributions, in a sample-efficient manner.
\item \textbf{Minimal redundancy:} The code should produce binary representations of minimal redundancy with regards to its probability estimate.
\item \textbf{High efficiency:} The code should have high coding efficiency, meaning, that encoding/decoding should have high throughput.
\end{itemize}

\subsubsection{CABAC}
\label{subsec: CABAC}

\begin{figure}[t]
\centering
\includegraphics[width=0.75\columnwidth,clip,keepaspectratio]{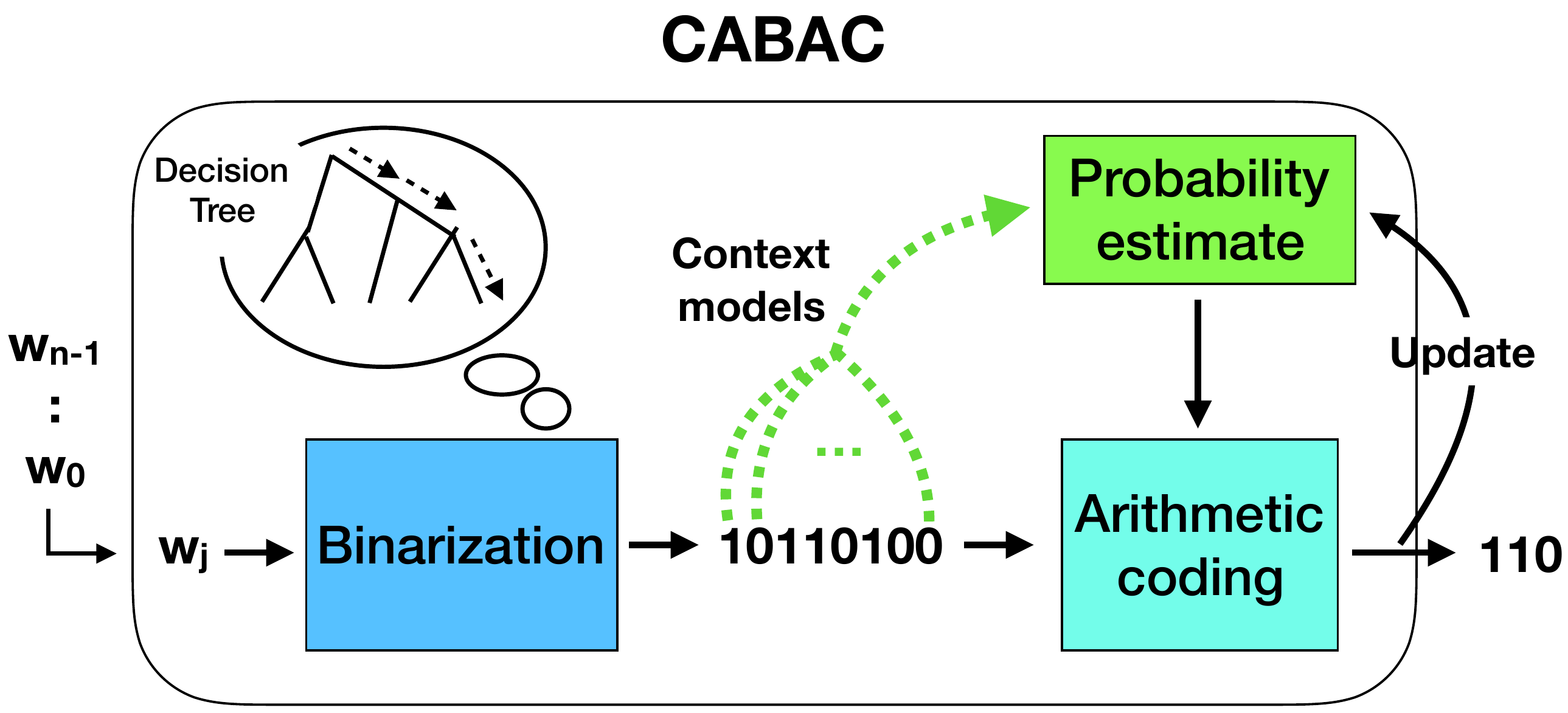}
\caption{Context-Adaptive Binary Arithmetic Coding (CABAC) is a universal lossless codec that encodes an $n$-long sequence of 1-dimensional values by: 1) representing each unique value by a binary string that corresponds to traversing a particular path on a predefined decision tree, 2) assigning to each decision (or bin) a probability model (or context models) and updating these upon encoding/decoding data, and 3) applying a binary arithmetic coder in order to encode/decode each bin.}
\label{Fig: cabac sketch}
\end{figure}

Context-based Adaptive Binary Arithmetic Coding is a form of universal lossless coding that fulfils all of the above  properties, in that it offers a high degree of adaptation, optimal code-lengths, and a highly efficient implementation. It was originally designed for the video compression standard H.264/AVC \cite{CABAC}, but it is also an integral part of its successor H.265/HEVC. It is well known to attain higher compression performance as well as higher throughput as compared to other entropy coding methods \cite{CABAC_efficient}. In short, it encodes each input sample by applying the following three stages:
\begin{enumerate}
\item \textbf{Binarization:} Firstly, it predefines a series of binary decisions (also called bins) under which each unique input sample element (or symbol) will be uniquely identified. In other words, it builds a predefined binary decision tree where each leaf identifies a unique input value.
\item \textbf{Context-modeling:} Then, it assigns a binary probability model to each bin (also named context model) which is updated on-the-fly by the local statistics of the data. This enables CABAC to model a high degree of different source distributions.
\item \textbf{Arithmetic coding:}  Finally, it employs an arithmetic coder in order to optimally and efficiently code each bin, based on the respective context model.
\end{enumerate}
Notice that, in contrast to two-part Huffman codes, CABACs encoder does \textbf{not need to encode its probability estimates}, since the decoder is able to analogously update its context models upon sequentially decoding the input samples. Codes that have this property are called \textit{backward-adaptive} codes. Moreover, its able to take \textbf{local correlations} into account, since the context models are updated by the local statistics of the incoming input data.

\subsection{Lossy coding}
\label{subsec: lossy coding}

In contrast to lossless coding, information is lost in the lossy coding process. This implies that the quantizer $Q$ is non invertible, and therefore $\exists w: D(w,q) \neq 0$.  An example of a distortion measure may be  the mean-squared error  $D(w,q) = ||w-q||_2^2$, but we stress that other measures can be considered as well (which will become apparent in section \ref{sec: DeepCABAC}).

The infimum of the rate-distortion objective \eqref{Eq: operational rate-distortion function} $\forall \lambda$ is referred to as the \textit{rate-distortion function}  in the source coding literature \cite{Wiegand_source_coding}, and it represents the fundamental bound on the performance of lossy source coding algorithms. However, as we have already discussed above, finding the most optimal code that follows the rate-distortion function is most often NP-hard, and can be calculated only for very few types and/or special cases of input sources. Therefore, in practice, we relax the problem until we formalize an objective that we can solve in a feasible manner.

Firstly,  we fix the binarization map $B$ by selecting a particular (universal) lossless code and condition the minimization of \eqref{Eq: operational rate-distortion function} on it. That is, now we only ask for the quantizer $Q$, along with its reconstruction values $Q^{-1}$, that minimize the respective rate-distortion objective. Secondly, we always assume that we encode an $n$-long 1-dimensional random process. Then, objective \eqref{Eq: operational rate-distortion function} simplifies to: given a lossless code $(B, B^{-1})$, find $(Q, Q^{-1})^* $ that
\begin{equation}
(Q, Q^{-1})^*  = \argmin{(Q, Q^{-1})} \mathbb{E}_{P(w_j)}[D(w_j,q_j) + \lambda L_Q(b_j)],
\label{Eq: scalar quantizer rate-distortion function}
\end{equation}
$\forall j\in \{0,...,n-1\}$, where $q_i\in \mathbb{Q}_s:= \{q_0,q_1,...,q_{K-1}\} \subset \mathbb{R}$ and $K = |\mathbb{Q}_s| < |\mathbb{W}_s|= n$.

For instance, if we choose $B$ such that it assigns a binary representation of fixed-length to all $w_j$, then the minimizer of \eqref{Eq: scalar quantizer rate-distortion function} can be found by applying the K-Means algorithm.

The minimizers of \eqref{Eq: scalar quantizer rate-distortion function} are called  \textit{scalar quantizers}, since they measure the distortion independently for each input sample. In contrast, \textit{vector quantizers} are those that result from minimizing \eqref{Eq: scalar quantizer rate-distortion function} when grouping a sequence of input samples together and measuring the distortion in the respective vector space. It is well known that the infimum of scalar codes are fundamentally more redundant than vector quantizers. Nevertheless, due to the associated complexity of vector quantizers, it is more common to apply scalar quantizers in practice. Moreover, the inherent redundancy of scalar quantizers is mostly negligible for most practical applications \cite{Wiegand_source_coding}.

We also want to stress that although the distortion in \eqref{Eq: scalar quantizer rate-distortion function} is measured independently for each sample, the binarization $b_j$  (and consequently the respective code-length) of each sample can still depend on the other samples by taking correlations into account.

\subsubsection{Scalar Lloyd algorithm} An example of an algorithm that finds a local optimum is the Lloyd algorithm. It approximates the average code-length of the quantized samples $q_j = (Q^{-1}\circ Q)(w_j)$ with the entropy of their empirical probability mass distribution (EPMD). Thus, it substitutes the code-length in \eqref{Eq: scalar quantizer rate-distortion function} by $L_C(b_j) = -\log_2{P_{\text{EPMD}}(q_j)}$ and applies a greedy algorithm in order to find the most optimal quantizer $Q$ and quantization points $Q^{-1}$ that minimize the respective objective. A pseudocode can be seen in the appendix (see algorithm \ref{algorithm:weighted-lloyds-algorithm}).

\subsubsection{CABAC-based RD-quantization} If we are given a set of quantization points $\mathbb{Q}_s$ and select CABAC as our posterior universal lossless code, then we can trivially minimize \eqref{Eq: scalar quantizer rate-distortion function} by sequentially quantizing the input samples. In the video coding standards, the set of quantization points are predefined by the particular choice of quantization strength $\lambda$ \cite{SzeBudSul_2014}. However, in the context of neural network compression we do not know of a good relationship between the quantization strength and the set of quantization points. In the next section \ref{sec: DeepCABAC} we describe how we tackled this problem.

\subsection{Model compression vs. source coding}
\label{subsec: model compression}
\begin{figure}[t]
\centering
\includegraphics[width=1\columnwidth,clip,keepaspectratio]{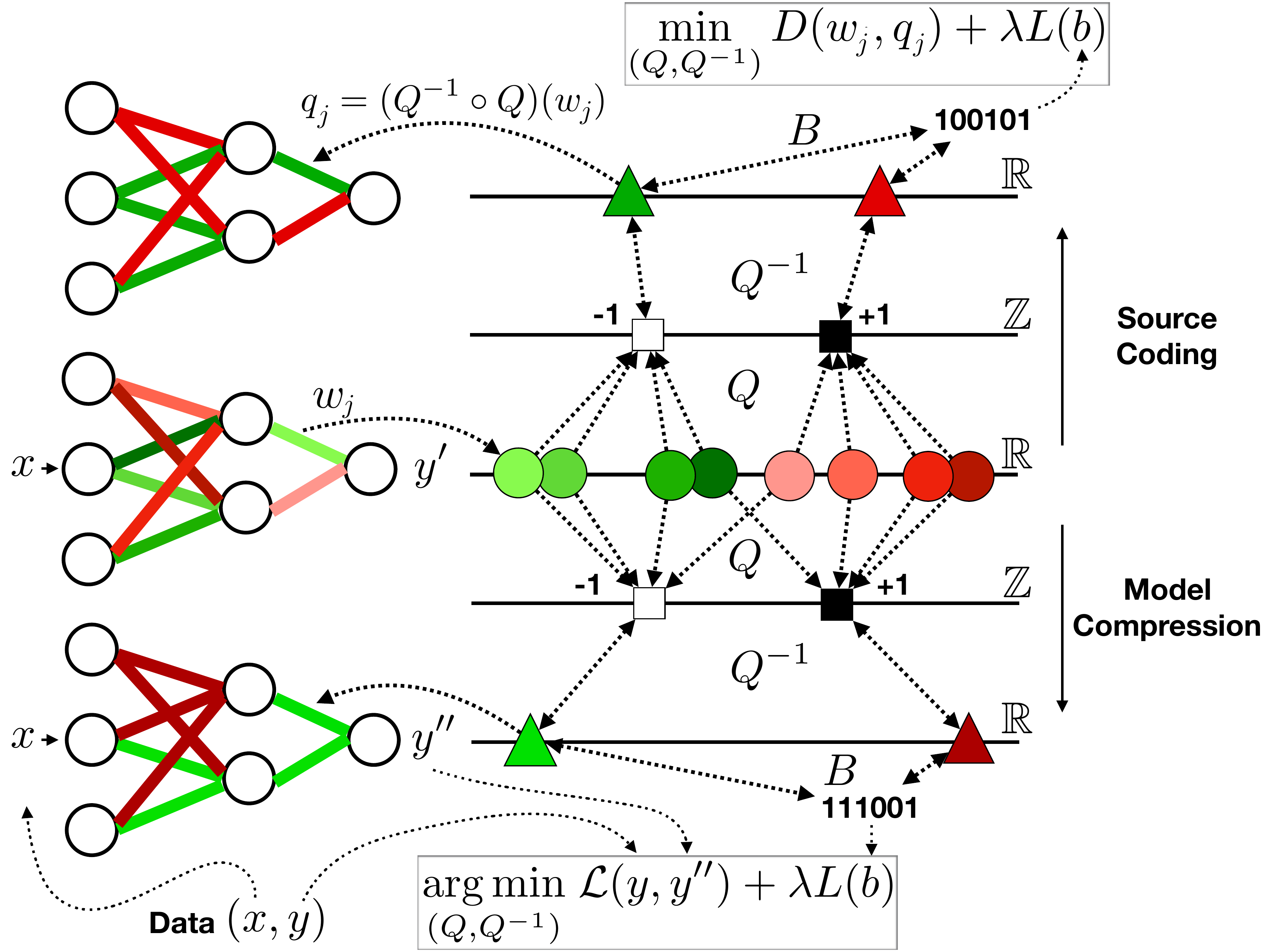}
\caption{Sketch displaying the fundamental difference between the source coding and the model compression problem. In the source coding problem, the parameters of a model are quantized by minimizing a rate-distortion function as in eq. \eqref{Eq: scalar quantizer rate-distortion function}, whereas in the model compression problem as in eq. \eqref{Eq: accuracy-rate quantization}. The main difference between the two paradigms lies in their measure of error, where the former is based on a distance measure between the unquantized and quantized parameters and the latter on the prediction performance of the quantized model. This results in different quantization schemes as solutions, which are displayed in the sketch. Different colors denote different parameter values. The different shapes correspond to different stages on the quantization procedure, with circles denoting the unquantized values, squares their respective integer representations and triangles their corresponding quantization points.}
\label{Fig: Quant problem}
\end{figure}

So far, we have reviewed some fundamental results of source coding theory. However, in this work, we are rather interested in the general topic of \textit{model compression}. There is a fundamental difference between both paradigms. Namely, now we are more interested in the predictive performance of the resulting quantized model rather than the distance between the quantized and original parameters. Figure \ref{Fig: Quant problem} highlights this distinction. We will now formalize the general model compression paradigm for the supervised learning setting. However, the problem can be analogously formulated for other learning tasks.

Firstly, we assume that we are given only one model sample with $n$ real-valued weight parameters (thus, here the input space is equivalent to the one discussed above). In addition, we assume a universal coding setting, where the decoder has no prior knowledge regarding the distribution of the parameter values. We argue that this simulates most real-world scenarios.

Let now $(x,y)\in \mathbb{D}$ be a set of data samples. Let further $y'\sim P(y'|x, w)$ denote the approximate posterior of  the data, parameterized by $w\in \mathbb{R}^n$. For instance, $P(y'|x, w)$ may be a trained neural network model with parameters $w$. Finally, let $B$ be a chosen and fix universal lossless code. Then, we aim to find a quantizer $Q^*$ that minimizes
\begin{equation}
(Q, Q^{-1})^* = \argmin{(Q, Q^{-1})} \; \sum_{(x,y)\in \mathbb{D}}\mathcal{L}(y, y'') + \lambda L_Q(b)
\label{Eq: accuracy-rate quantization}
\end{equation}
with $y''\sim P(y''|x,q)$ being outputs of the quantized model $q = (Q^{-1}\circ Q)(w)$ and $b = (B\circ Q)(w)$.

The first term in \eqref{Eq: accuracy-rate quantization} expresses the minimization of the usual learning task of interest, whereas the second term explicitly expresses the code-length of the model. This minimization objective is well motivated from the Minimum Description Principle (MDL) \cite{MDL}. However, finding the minimum of \eqref{Eq: accuracy-rate quantization} is also most often NP-hard. This motivates further approximations where, as a result, one can directly apply techniques from the source coding literature in order to minimize the desired objective.

\subsubsection{Relaxation of the model compression problem into a source coding problem}
We may further assume that the given unquantized model has been pre-trained on the desired task and that it reaches satisfactory accuracies. Then, in such cases, it is reasonable to replace the first term in \eqref{Eq: accuracy-rate quantization} by the KL-Divergence between the unquantized model $P(y'|x, w)$ and the respective quantized model $P(y''|x, q)$. That is, now we want to quantize our model such that its output distribution does not differ too much from its original version.

Furthermore, if we now assume that the output distributions do not differ too much from each other, then we can approximate the KL-Divergence with the Fisher Information Matrix (FIM). Concretely,
\begin{equation}
\mathbb{E}_{P_{\mathbb{D}}} [D_{KL}(y''||y')] = \delta w F \delta w^T + \mathcal{O}(\delta w^2)
\label{Eq: Fisher}
\end{equation}
with $\delta w = q-w$ and
\[F:= \mathbb{E}_{P_{\mathbb{D}}} \mathbb{E}_{P(y'|x,w)}[\partial_w\log P(y'|x,w)(\partial_w\log P(y'|x,w))^T]\]

Then, by substituting \eqref{Eq: Fisher} in \eqref{Eq: accuracy-rate quantization} we get the following minimization objective
\begin{equation}
(Q, Q^{-1})^* = \min_{(Q, Q^{-1})} \; (q-w) F (q-w)^T + \lambda L_Q(b)
\label{Eq: accuracy-rate function fisher approx}
\end{equation}
Objective \eqref{Eq: accuracy-rate function fisher approx} now follows the same paradigm as the usual source coding problem. However, with the peculiarity that now $D(w,q)$ (approximately) measures the distortion of $w$ and $q$ in the space of output distributions instead the Euclidian space. The advantage of the rate-distortion objective \eqref{Eq: accuracy-rate function fisher approx} is that,  after the FIM has been calculated, it can be solved by applying common techniques from the source coding literature, such as the scalar Lloyd algorithm.

However, minimizing \eqref{Eq: accuracy-rate function fisher approx} as well as estimating the FIM for deep neural networks usually requires considerable computational resources, and is most often infeasible for practical applications. Therefore,  we further approximate it by only its diagonal elements (FIM-diagonals), which can be efficiently estimated (see appendix). As a result, \eqref{Eq: accuracy-rate function fisher approx} simplifies into
\begin{equation}
(Q, Q^{-1})^*= \argmin{(Q, Q^{-1})}\; F_i(q_i-w_i)^2 + \lambda L_Q(b)
\label{Eq: scalar diag fisher rate-distortion function}
\end{equation}
$\forall i\in \{0,...,n-1\}$, which can be feasibly solved.

In the next section we will give a thorough description of our proposed coder. Its design complies with all desired properties that a coder for neural network compression should have.

\section{DeepCABAC}
\label{sec: DeepCABAC}

\begin{figure}[t]
\centering
\includegraphics[width=1\columnwidth,clip,keepaspectratio]{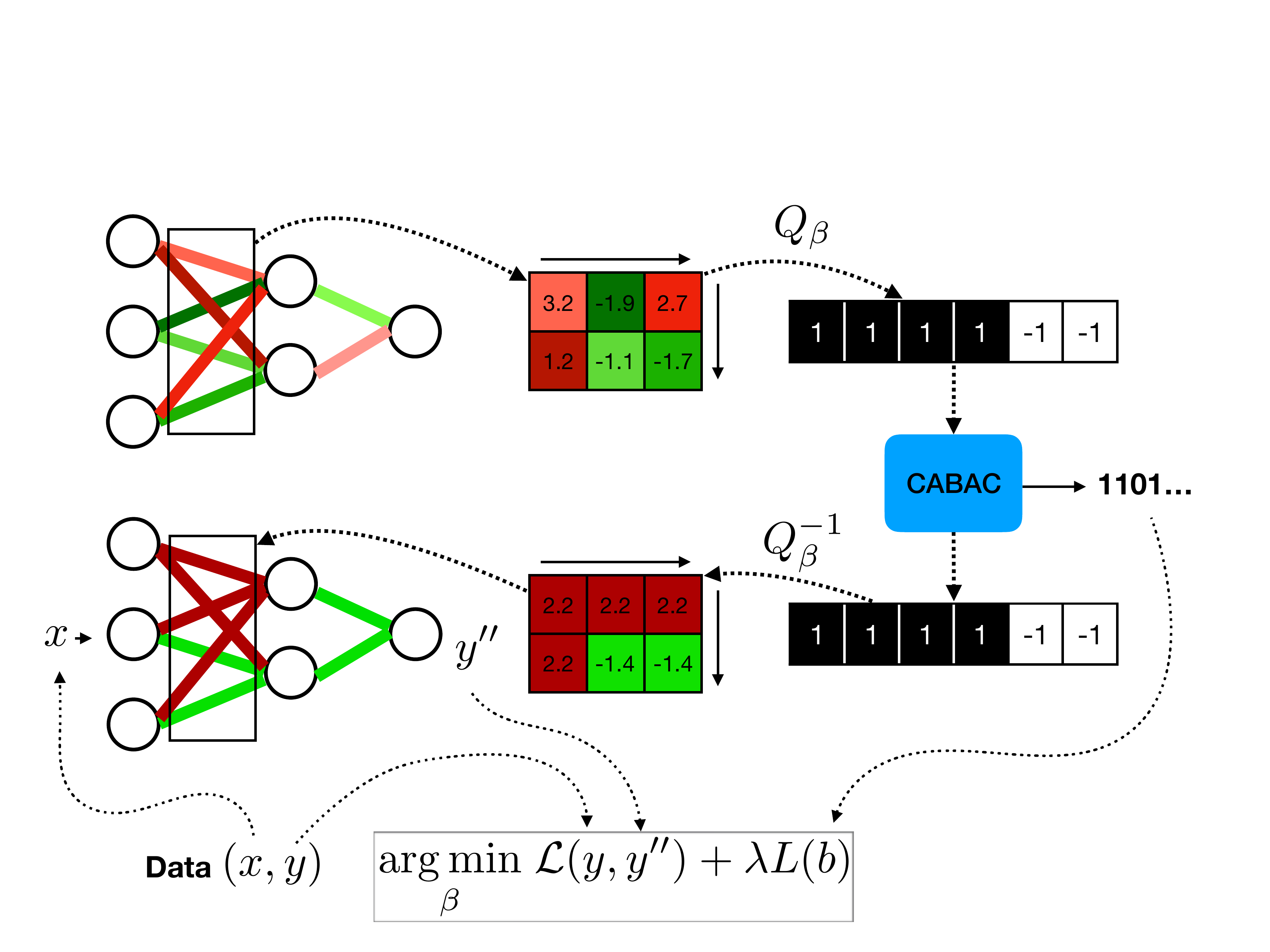}
\caption{Sketch of the DeepCABAC compression procedure. Firstly, DeepCABAC scans the weight parameters of each layer of the network in row-major order. Then, it selects a particular hyperparameter $\beta$ that will define the set of quantization points. Subsequently, it applies a quantizer on to the weight values that minimizes the respective weighted rate-distortion function \eqref{Eq: deepcabac quantization}. Then, it compresses the quantized parameters by applying our adapted version of CABAC. Finally, it reconstructs the network and measures the respective accuracy of it. The process is repeated for different hyperparameters $\beta$ until the desired trade-off between accuracy and bit-size of the network is achieved.}
\label{Fig: DeepCABAC}
\end{figure}

In light of the discussion in the previous section, we can highlight a set of desiderata that a coder for neural network compression should have.
\begin{itemize}
\item \textbf{Minimal redundancy}: State-of-the-art deep neural networks usually contain millions of parameters. Thus, any type of redundancy in the weight parameters may imply several additional MB being stored. Hence, the code should output a binary representation with  minimal redundancy per weight element.
\item \textbf{Universality:} The code should be applicable to any type of incoming neural networks, without having to know their distribution a priori. Hence, the code should entail a mechanism that allows it to adapt to a rich number of possible parameter distributions.
\item \textbf{High coding efficiency:} The computational complexity of encoding/decoding should be minimal. In particular, the throughput of the decoder should be very high if performing inference on the compressed representation is desired.
\item \textbf{Configurable error vs. compression strength:} The coder should have a hyperparameter that controls the trade-off between the compression strength and the incurred prediction error.
\item \textbf{High data efficiency:} Minimizing \eqref{Eq: accuracy-rate quantization} implies access to data. Hence, it is desirable that the coder finds a (local) solution with the least amount of data samples possible.
\end{itemize}

\subsection{DeepCABAC's coding procedure}
We propose a coding algorithm that satisfies all the above properties. We named it \textit{DeepCABAC}, since it's based on applying CABAC on the networks quantized weight parameters. Figure \ref{Fig: DeepCABAC} shows the respective compression scheme. It performs the following steps:
\begin{enumerate}
\item it extracts the weight parameters of the neural networks layer-by-layer in row-major order\footnote{Thus, it assumes a matrix form where the parameters are scanned from left-to-right, top-to-bottom.}.
\item Then, it selects a particular value $\beta$ which defines the set of quantization points.
\item Subsequently, it quantizes the weight values by minimizing a weighted rate-distortion function, which implicitly takes the impact of quantization on the accuracy of the network into account.
\item Then, it compresses them by applying our adapted version of CABAC.
\item Finally, it reconstructs the network and evaluates the prediction performance of the quantized model.
\item The process is repeated for a set of hyperparameters $\beta$, until the desired accuracy-vs-size trade-off is achieved.
\end{enumerate}

This approach has several advantageous properties. Firstly, it applies CABAC to the quantized parameters and therefore we ensure that the code satisfies the desiderata 1-3. Secondly, by conducting the compression for a set of hyperparameters for the quantizer we can select the desired pareto-optimal solutions of the accuracy vs. bit-size plane, thus satisfying property 4. Finally, since only one evaluation of the model is required in the process, a significantly lower amount of data samples is required for the compression process than usually employed for training.

In the following we will explain in more detail the different components of DeepCABAC.

\subsection{Lossless coder of DeepCABAC}
\begin{figure}[t]
\centering
\includegraphics[width=0.85\columnwidth,clip,keepaspectratio]{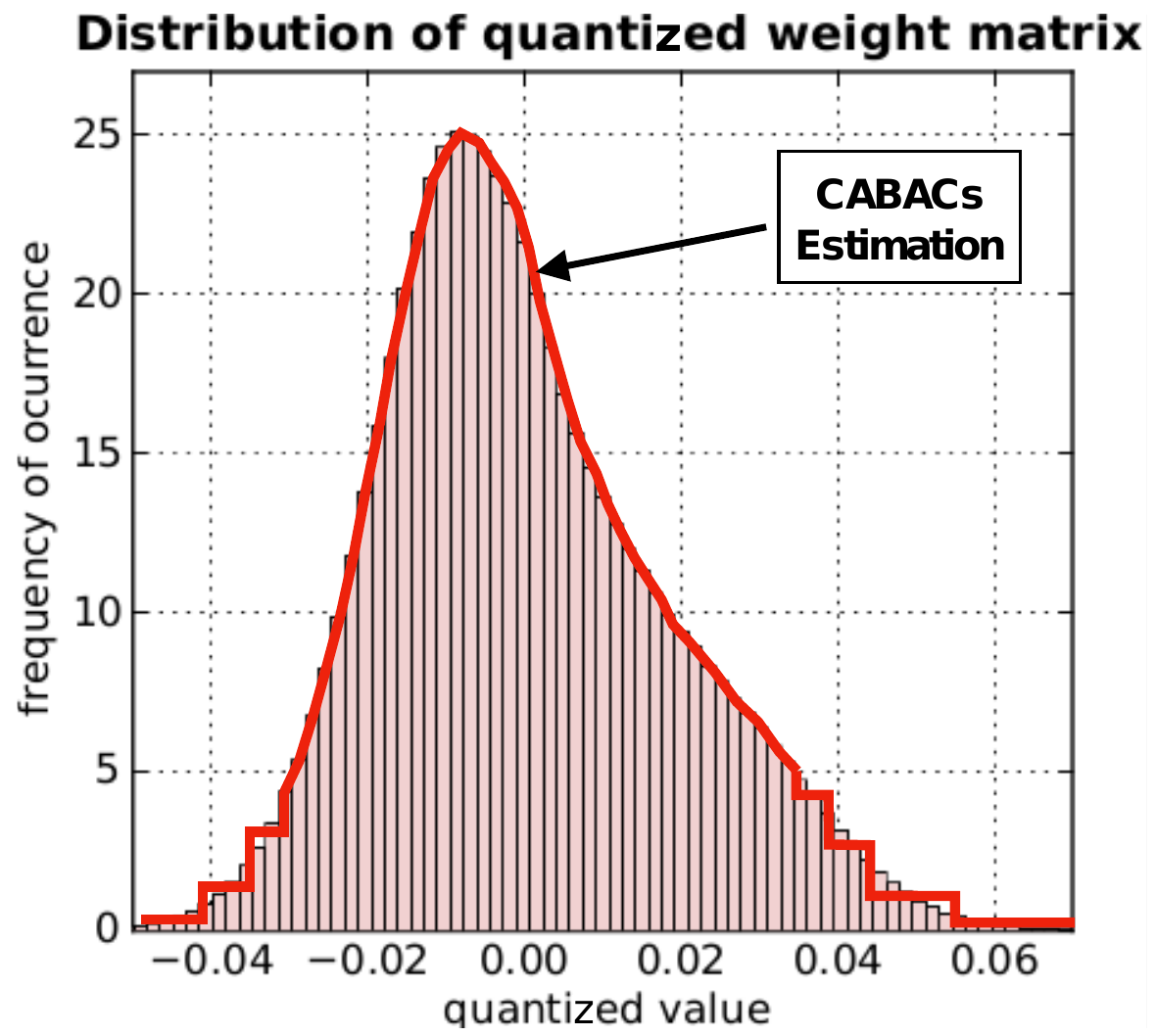}
\caption{Distribution of the weight matrix of the last layer of VGG-16 (trained on ImageNet) after uniform quantization over the range of values. In red is CABACs (possible) estimation of the distribution. The first $n+2$ bits allow to adapt to any type of shape around 0 since they are encoded with regards to a context model. The remainder can only approximate the shape by a step-like distribution, since they are encoded with an Exponential-Golomb where the fixed-length parts are encoded without a context model.}
\label{Fig: VGG class distr}
\end{figure}

Consider the weight distribution of the last fully-connected layer of the trained VGG16 model displayed in figure \ref{Fig: VGG class distr}. As we can see, there is \textbf{one peak near 0} and the distribution is \textbf{asymmetric} and \textbf{monotonically decreasing} on both sides. In our experience, all layers we have studied so far have weight distributions with similar properties. Hence, in order to accommodate to this type of distributions, we adopted the following binarization procedure.

Given a quantized weight tensor in its matrix form\footnote{For fully-connected layers this is trivial. For convolutional layers we converted them into their respective matrix form according to \cite{NVIDIA_convolution}.}, DeepCABAC scans the weight elements in row-major order and binarizes them as follows:
\begin{center}
\begin{tikzpicture}
\filldraw[fill=black!20!white, draw=black] (0,0) rectangle (1.25,0.45) node[pos=0.5] {\small SigFlag};
\filldraw[fill=black!20!white, draw=black] (1.25,0) rectangle (2.75,0.45) node[pos=0.5] {\small SignFlag};
\filldraw[fill=black!20!white, draw=black] (2.75,0) rectangle (5,0.45) node[pos=0.5] {\small AbsGr($n$)Flags};
\draw (5,0.45) rectangle (7.5,0.8) node[pos=0.5] {\footnotesize Exp.-Golomb};
\filldraw[fill=black!20!white, draw=black]  (5,0.0) rectangle (6.25,0.45) node[pos=0.5] {\small Unary};
\filldraw[fill=blue!30!white, draw=black]  (6.25,0.0) rectangle (7.5,0.45) node[pos=0.5] {\small FL};
\end{tikzpicture}
\end{center}

\begin{enumerate}
\item The first bit, \textit{sigFlag}, determines if the weight element is a significant element or not. That is, it indicates if the weight value is 0 or not. This bit is then encoded using a binary arithmetic coder, according to its respective context model (color-coded in grey). The context model is initially set to 0.5 (thus, 50\% probability that a weight element is 0 or not), but will automatically be updated to the local statistics of the weight parameters as DeepCABAC encodes more elements.
\item Then, if the element is not 0, the sign bit or  \textit{signFlag} is analogously encoded, according to its respective context model.
\item Subsequently, a series of bits are analogously encoded, which determine if the element is greater than $1,2,...,n \in \mathbb{N}$ (hence \textit{AbsGr($n$)Flags}). The number $n$ becomes a hyperparameter for the encoder.
\item Finally, the remainder is encoded using an Exponential-Golomb\footnote{To recall, the Exponential-Golomb code encodes a positive integer $2^k <i \leq 2^{k+1}$ by firstly encoding the exponent $k$ using an unary code and subsequently the remainder $r = i-2^k$ in fixed-point representation.} code\cite{exp_goulomb_code}, where each bit of the unary part are also encoded relative to their context-models. Only the fixed-length part of the code is not encoded using a context-model (color-coded in blue).
\end{enumerate}

\begin{figure}[t]
\centering
\includegraphics[width=0.75\columnwidth,clip,keepaspectratio]{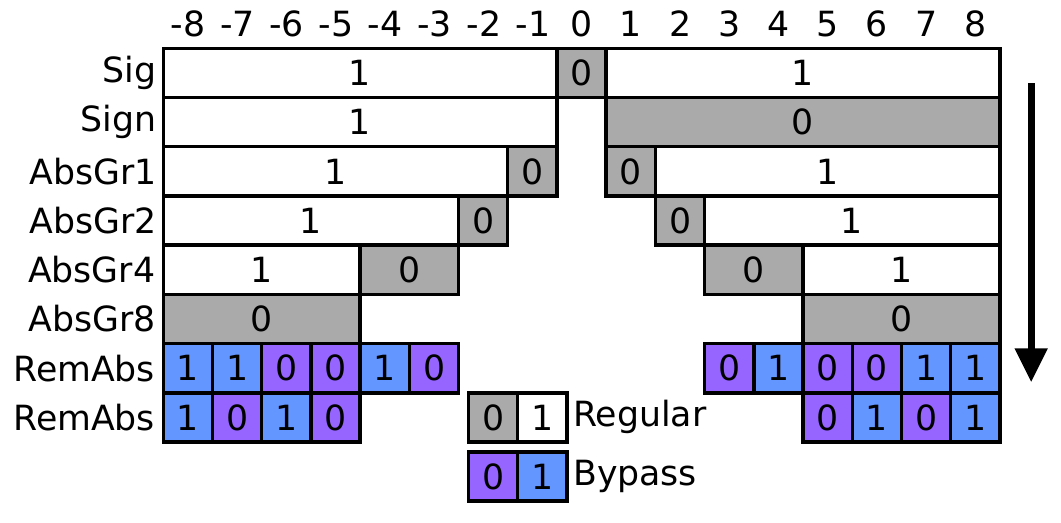}
\caption{ DeepCABAC binarization of neural networks. It encodes each weight element by performing the following steps: 1) encodes a bit named \textit{sigflag} which determines if the weight is a significant element or not (in other words, if its 0 or not). 2) If it's not 0, then the sign bit, \textit{signflag}, is encoded. 3) Subsequently, a series of bits are encoded, which indicate if the weight value is greater equal than $1, 2, ..., n \in \mathbb{N}$ (the so called  \textit{AbsGr(n)Flag}). 4) Finally, the remainder is encoded. The gray bits (also named regular bins) represent bits that are encoded using an arithmetic coder according to a context model. The other bits, the so called bypass bins, are encoded in fixed-point form. For instance, in the above diagram $n=1$, and therefore  $1 \rightarrow$ \textcolor{gray}{100} , $-4\rightarrow$ \textcolor{gray}{11110}\textcolor{blue}{1} or $7\rightarrow$ \textcolor{gray}{101110}\textcolor{blue}{10} .}
\label{Fig: deepcabac}
\end{figure}

For instance, assume that $n=1$, then the integer $-4$ would be represented as \textcolor{gray}{11110}\textcolor{blue}{1}, or the 7 as \textcolor{gray}{101110}\textcolor{blue}{10}. Figure \ref{Fig: deepcabac} depicts an example scheme of the binarization procedure.

The first three parts of the binarization scheme empower CABAC to adapt its probability estimates to any shape distribution around the value 0 and, therefore, to encode the most frequent values with minimal redundancy. For the remainder values, we opted for the Exponential-Golomb code since it automatically assigns smaller code-lengths to smaller integer values. However, in order to further enhance its adaptability, we also encode its unary part with the help of context models. We left the fixed-length part of the Golomb code without context models, meaning that we approximate the distribution of those values by a uniform distribution (see figure \ref{Fig: VGG class distr}). We argue that this is reasonable since usually the distribution of large numbers become more and more flat, and it comes with the direct benefit of increasing the efficiency of the coder.

\subsection{Lossy coder of DeepCABAC}
After establishing CABAC as our choice of universal lossless code, now we aim to find the optimal quantizer that minimizes the objective stated in \eqref{Eq: accuracy-rate quantization} (section \ref{subsec: model compression}). To recall, this involves the optimization of two components (see figure \ref{Fig: Quant problem}):
\begin{itemize}
\item \textbf{Assignment:} finding the quantizer $Q$ that assigns the optimal set of integers to each weight parameter
\item \textbf{Quant. points:} finding the optimal quantization points $q_j = (Q^{-1}\circ Q)(w_j)$.
\end{itemize}
Since neural networks usually rely on scalable, gradient-based minimization techniques in order to optimize their loss function, finding the quantizers that solve \eqref{Eq: accuracy-rate quantization} becomes infeasible in most cases since $Q$ is a non-differentiable map. Therefore, we opted for a simpler approach.

Firstly, we decouple the assignment map $Q$ and the  quantization points $Q^{-1}$ from each other and optimize them independently. The quantization points then become hyperparameters for the quantizer, and their values are selected such that they minimize the loss function directly. This separation between $Q$ and $Q^{-1}$ was empirically motivated, since we discovered that the networks performance is significantly more sensitive to the choice of $Q^{-1}$ than to the assignment $Q$. We discuss this in more detail in the experimental section.

\subsubsection{The quantization points}
Since finding the correct map $Q^{-1}$ for a large number of points can be very complex, we constrain them to be equidistant to each other with a specific step-size $\Delta$. That is, each point $q_k$ can be rewritten as to be $q_k = \Delta I_k$ with $I_k \in \mathbb{Z}$. This does not only considerably simplify the problem, but it does also encourage fixed-point representations which can be exploited in order to perform inference with lower complexity \cite{QNNPACK, TFlite}.

\subsubsection{The assignment}
Hence, the quantizer has two configurable hyperparameters $\beta = (\Delta, \lambda)$, the former defining the set of quantization points and the latter the quantization strength. Once a particular tuple is given, the quantizer $Q_{\beta}$ will then assign each weight parameter $w_i$ to its corresponding quantization point by minimizing the weighted rate-distortion function
\begin{equation}
Q_{\beta}(w_i) = k^* = \argmin{k} \; F_i(w_i-q_k)^2 + \lambda L_{ik}
\label{Eq: deepcabac quantization}
\end{equation}
$\forall i\in \{0,...,n-1\}$, where $L_{ik}$ is the code-length of the quantization point $q_k$ at the weight $w_i$ as estimated by CABAC.

As previously mentioned, we perform a grid-search algorithm over the hyperparameters $\Delta$ and $\lambda$ in order to find the quantizer configuration that achieves the desired accuracy vs. bit-size trade-off. However, for that we need to define a predefined set of hyperparameter candidates to look for, in particular for the step-sizes $\Delta$. In this work we considered two approaches for finding the set of step-sizes, which we denote as \textit{DeepCABAC-version1} (DC-v1) and \textit{DeepCABAC-version2} (DC-v2).

\subsubsection{DeepCABAC-version1 (DC-v1) }
In DC-v1 we firstly estimate the diagonals of the FIM by applying scalable Bayesian techniques. Concretely,  we  parametrize the network with a fully-factorized gaussian posterior and minimize the variational objective proposed in  \cite{VD_sparsifies}. As a result, we obtain a mean $\mu_j$ and a standard deviation $\sigma_j$ for each parameter, where the former can be interpreted as its (new) value (thus $w_i \rightarrow \mu_i$) and the latter as a measure of their ``robustness'' against perturbations. Therefore, we simply replaced $F_i = 1/\sigma_i^2$ in \eqref{Eq: deepcabac quantization}. This is also well motivated theoretically since \cite{CLP_achile} showed that the variance of the parameters approximate the diagonals of the FIM for a similar variational objective. We also provide a more thorough discussion and a precise connection between them in the appendix.

After the FIM-diagonals have been estimated, we define the set of considered step-sizes as follows:
\begin{equation}
q_{k} = \Delta I_k, \quad \Delta =\frac{2|w_{\max}|}{ \frac{2|w_{\max}|}{\sigma_{\min}} + S},  \quad S,I_k \in \mathbb{Z}
\label{Eq: quantization points}
\end{equation}
where $\sigma_{\min}$ is the smallest standard deviation and $w_{\max}$ the parameter with highest magnitude value. $S$ is then the quantizers hyperparameter, which controls the ``coarseness'' of the quantization points. By selecting $\Delta$ in such a manner we ensure that the quantization points lie within the range of the standard deviation of each weight parameter, in particular for values $S \geq 0$. Hence, we selected $S$ to be $S\in \{0,1,...,256\}$.

One advantage of this approach is that we can have one global hyperparameter $S$ for the entire network, but each layer will still attain a different value for its step-size if we select one $\sigma_{\min}$ per layer. Thus, with this approach we can adapt the step-size to the layer's sensitivity to perturbations. Moreover, the quantization will also take the sensitivity of each single parameter into account.

\subsubsection{DeepCABAC-version2 (DC-v2)}
Estimating the diagonals of the FIM can still be computationally expensive since it requires the application of the backpropagation algorithm for several iterations in order to minimize the variational objective.  Moreover, it only offers an approximation of the robustness of each parameter, and can therefore sometimes be misleading and limit the potential compression gains that can be achieved. Therefore, due to simplicity and complexity reasons, we also considered to directly try to find a good set of candidates $\Delta \in \{\Delta_0,...,\Delta_{m-1}\}$. We do so by applying a first round of the grid-search algorithm while applying a nearest-neighbor quantization scheme (that is, for $\lambda=0$).  This allows us to identify the range of step-sizes that do not considerably harm the networks accuracy when applying the simplest quantization procedure. Then, we quantize the parameters as in eq. \eqref{Eq: deepcabac quantization}, but without the diagonals of the FIM (thus, $F_j = 1 \; \forall j$).

Under a limited computational budget, this approach has the advantage that we can directly search for a more optimal set of step-sizes $\Delta$ since we spare the computational complexity of having to estimate the FIM-diagonals. However, since DC-v2 considers only one global step-size for the entire network, it cannot adapt to the different sensitivities of the layers.

\section{Related work}
\label{sec: related work}
There has been a plethora of work focusing on  the topic of neural network compression \cite{DLC_survey, DLC_survey2}. In some way or another, all of them try to partially solve the general model compression objective \eqref{Eq: accuracy-rate quantization} (section \ref{subsec: model compression}).  In the following we will thoroughly examine the currently proposed approaches and discuss some of their advantages and disadvantages.

\subsection{Lossy neural network compression}
Some of the insofar proposed approaches include:

\subsubsection{Trained scalar quantization}
These are methods that aim to minimize the model compression objective \eqref{Eq: accuracy-rate quantization} by applying training algorithms that learn the optimal quantization map and reconstruction values. Inter alia, this includes sparsification methods \cite{Opt_brain_surgeon, Opt_brain_damage, Learning_Weight_and_Connections, dynamic_network_surgery, VD_sparsifies, l0_regularization, NIPS2018_7644} which try to minimize the $L_0$-norm of the network's parameters. Others attempted to find optimal binary or ternary weighted networks \cite{XNOR_net, BinaryNet, trained_terniary_quant, Terniray_weights}, or a more general set of (locally) optimal quantizers \cite{ deep_compression, soft_weight_sharing, BayesianCompression, VariationalNQ, Entropy_constrained_DNN}.

Although these methods are able to attain very high compression ratios, they are also very computationally expensive in that several training iterations over a large training set have to be performed in order to find the optimal quantized network. In contrast, DeepCABAC does \textbf{not require} any \textbf{retraining}, nor access to the full training dataset in order to be applicable.

 \subsubsection{Non-trained scalar quantization methods}
Another line of work has focused on implicitly minimizing \eqref{Eq: accuracy-rate quantization}. They also rely on distance measures for quantizing the network's parameters \cite{weighted_entropy_quantization_DNN, DNN_compression_FIM_diag, towards_limit_quantization_DNN, Universal_dnn_compression}. In fact, these methods can be seen as special cases of \eqref{Eq: scalar diag fisher rate-distortion function}, in that they either use different approximations of the FIM-diagonals or apply other minimization algorithms. To the best of our knowledge, mainly two quantizers are widely applied by the community, either the scalar uniform quantizers or the weighted scalar Lloyd algorithm. Usually, the former basically consist on uniformly spreading $K\in \mathbb{N}$ quantization points over the range of parameter values and then applying  nearest-neighbor quantization on to them \cite{towards_limit_quantization_DNN, deep_compression, BayesianCompression}. The latter consists of applying the scalar Lloyd algorithm in order to find the most optimal quantizer that minimizes a weighted rate-distortion objective \eqref{Eq: scalar diag fisher rate-distortion function}. In particular,\cite{towards_limit_quantization_DNN} considers the diagonal elements of the empirical average of the Hessian of the loss function, which has a close connection to the FIM-diagonals (see appendix for a comprehensive discussion).

Applying quantization methods that do not rely on retraining are significantly less computationally expensive. But their compression gains are heavily limited by the networks unquantized parameter distribution, since they rely on a distance measure for quantization. Moreover, as already mentioned, most of these methods do only implicitly take the impact on to the accuracy of the network into account. In contrast, DeepCABAC does explicitly take the accuracy of the network into account since its hyperparameters are optimized with regards to it.

%\subsubsection{Vector quantization} \textcolor{red}{Insert citations}
%Some have proposed to apply vector quantization methods on to the networks parameters \cite{compressing_dnn_VQ, Universal_dnn_compression}. Others have proposed matrix decomposition methods for the weight matrices, such as low-rank decompositions \cite{} or SVD methods \cite{}, which can be also seen as special cases of vector quantization methods. Whilst these coders may in principle produce less redundant code-lengths, usually they are limited by their coding complexity \cite{Wiegand_source_coding}.

\subsection{Lossless neural network compression}
In the field of lossless network compression, we are given an already quantized model and we want to apply a universal lossless code to its parameters in order to maximally compress it. Hence, this setting is entirely equivalent to the usual lossless source coding setting discussed in sections \ref{subsec: lossless coding} and \ref{subsec: universal coding}, and therefore all of its theorems and results can be applied in a straight-forward manner. Nevertheless, most of the previous work did not apply state-of-the-art universal lossless compression algorithms to them. Instead, these are some of the most commonly used:

\subsubsection{Fixed-length numerical representations}
These methods reduce the bit-length representation of the parameter values after quantization \cite{XNOR_net, BinaryNet, trained_terniary_quant, Terniray_weights, DBLP:journals/corr/MellempudiKDMK17, FP_DNN_IBM, fxpnet, incremental_NN_quant_intel, 8bit_training_intel, 8bit_training_IBM}. They usually have the advantage of immediately reducing the complexity for performing inference, however, at the expense of having a highly redundant network representation.

\subsubsection{Scalar Huffman code}
Others applied the scalar Huffman code on to quantized neural networks \cite{towards_limit_quantization_DNN, Universal_dnn_compression}. However, as we have already discussed in section \ref{subsec: lossless coding}, this code has several disadvantages compared to other state-of-the-art lossless codes such as arithmetic codes. Probably the most prominent one is that this code is suboptimal in that it incurs up to 1 bit of redundancy per parameter being encoded. This can be quite significant for large networks with millions of parameters. For instance, VGG16 \cite{VGG} contains 138 million parameters, meaning that the binary representation of any quantized version of it may have about \textbf{17MB} of redundancy if we encode it using the scalar Huffman code.

\subsubsection{Compressed matrix representations}
Most of the literature that sparsify the networks parameters aim to convert the resulting networks into a compressed sparse matrix representation, e.g., the \textit{Compressed Sparse Row} (CSR) representation. These matrix data structures do not only offer compression gains, but also an efficient execution of the associated dot product algorithm \cite{sparse_matrices_research}. Similarly, \cite{Simon_lossless_dnn_compression1} proposed  two novel matrix representation, the \textit{Compressed Entropy Row} (CER) and \textit{Compressed Shared Elements Row} (CSER) representations, that are provably more optimal than the CSR with regards to both, compression and execution efficiency when the networks parameters have low entropy statistics.

However, these matrix representations are also redundant in that they do not approach the reachable entropy limit \eqref{Eq: Minimal average code-length} (section \ref{subsec: lossless coding}). \cite{deep_compression} attempted to extract some of the redundancies entailed in the CSR  representations by applying a scalar Huffman code to its numerical arrays. However, this has again the same limitations that come by applying the scalar Huffman code.

\subsection{Compression pipelines/frameworks}
Among all different proposed approaches for deep neural network compression there is one paradigm that stands out in that very high compression gain can be achieved with it \cite{deep_compression, VD_sparsifies, BayesianCompression, improved_BayesianCompression, Entropy_constrained_DNN}. Namely,  it consists on applying four different compression stages:
\begin{enumerate}
\item \textbf{Sparsification:} Firstly, the networks are maximally sparsified by applying a trained sparsification technique.
\item \textbf{Quantization:} Then, the non-zero elements are quantized by applying one of the non-trained quantization techniques.
\item \textbf{Fine-tuning:} Subsequently, the quantization points are fine-tuned in order to recover the accuracy loss incurred by the quantization procedure.
\item \textbf{Lossless compression:} Finally, the quantized values are encoded using a lossless coding algorithm.
\end{enumerate}
Hence, DeepCABAC is designed to enhance points 2 and 4. As we will see in the next section, DeepCABAC is able to considerably boost the attainable compression gains, surpassing previously proposed methods for steps 2 and 4.

\section{Experiments}
\label{sec: experiments}
In this section we benchmark DeepCABAC and compare it to other compression algorithms. We also design further experiments with the purpose to shed light on the effectiveness of its different components.

\subsection{General compression benchmark}
\label{subsec: Exp1}

\begin{table}[!t]
%% increase table row spacing, adjust to taste
\centering
\caption{Compression ratios achieved at no loss of accuracy when applying different coding methods. \textit{DC-v1} \& \textit{DC-v2} denote the two versions of DeepCABAC, whereas \textit{Lloyd} denotes the weighted Lloyd algorithm and \textit{uniform} the nearest-neighbor quantization scheme. For the latter two, we report the best compression results attained after applying scalar Huffman, CSR-Huffman \cite{deep_compression} and the bzip2 lossless coding algorithms on to the quantized networks. In parenthesis are the resulting top-1 accuracies, and the sparsity ratios achieved as measured $\frac{|w\neq 0|}{|w|}$.}
\vspace{0.1in}

\resizebox{\columnwidth}{!}{%
\begin{tabular}{M{1.5cm}M{1.0cm}M{1.0cm}|M{1.0cm}M{1.0cm}M{1.0cm}M{1.cm}}
\toprule
\thead{\normalsize Models \\ (Spars.  [\%])} \vspace{0.5mm}  & \thead{\normalsize Org. acc. \\ (top1  [\%])}  & \thead{\normalsize Org. \\ \normalsize size} & \thead{\normalsize DC-v1} [\%] & \thead{\normalsize DC-v2} [\%] & \thead{\normalsize Lloyd} [\%] & \thead{\normalsize Uniform} [\%] \\
\bottomrule
\multicolumn{7}{c}{\thead{\normalsize NON-SPARSE } } \\%\thead{\normalsize - \\  (-)}

\thead{\normalsize VGG16} & \normalsize 69.94 & \thead{\normalsize 553.43 \\ MB}  & \thead{\normalsize 5.84 \\  (69.44)} &  \thead{\normalsize \textbf{3.96} \\  (69.54)} & \thead{\normalsize 7.74 \\  (69.50)}  & \thead{\normalsize 17.37 \\  (69.90)}  \\%
\midrule
\thead{\normalsize ResNet50} & \normalsize 74.98 & \thead{\normalsize 102.23 \\ MB}  & \thead{\normalsize \textbf{10.14} \\  (74.40)} & \thead{\normalsize \textbf{10.14} \\  (74.51)} & \thead{\normalsize 13.04\\ (74.74)} & \thead{\normalsize 15.58\\ (74.64)}  \\
\midrule
\thead{\normalsize MobileNet\\ \normalsize -v1} & \normalsize 70.69 & \thead{\normalsize 17.02\\ MB}  & \thead{\normalsize \textbf{21.40}\\ (70.21)}  & \thead{\normalsize 22.08\\ (70.21)}  & \thead{\normalsize 15.00\footnotemark{}\\ (68.10)} & \thead{\normalsize 24.23\\ (70.10)}  \\
\midrule
\thead{\normalsize Small-\\ \normalsize VGG16} & \normalsize 91.54 & \thead{\normalsize 60.01\\ MB} & \thead{\normalsize 6.35\\ (91.11)} & \thead{\normalsize \textbf{5.88}\\ (91.13)} & \thead{\normalsize 9.98\\ (91.59)} &  \thead{\normalsize 16.18 \\  (91.53) } \\
\midrule
\thead{\normalsize LeNet5} & \normalsize 99.46 & \thead{\normalsize 1722 \\ KB}  & \thead{\normalsize 3.77 \\   (99.23)} & \thead{\normalsize \textbf{2.52} \\  (99.12)} &  \thead{\normalsize 3.96\\ (98.96)} & \thead{\normalsize 20.60 \\ (99.45)}  \\
\midrule
\thead{\normalsize LeNet-300\\ \normalsize -100} & \normalsize 98.32 & \thead{\normalsize 1066\\ KB}  & \thead{\normalsize 8.61 \\  (98.04)} & \thead{\normalsize \textbf{5.87} \\  (98.00)} & \thead{\normalsize 8.07\\ (97.92)} & \thead{\normalsize 15.01 \\ (98.30)}  \\
\bottomrule
  \multicolumn{7}{c}{\thead{\normalsize SPARSE } } \\

\thead{\normalsize VGG16 \\ \small (9.85)} & \normalsize 69.43 & \thead{\normalsize 553.43\\ MB}  & \thead{\normalsize \textbf{1.58} \\  (69.43)} & \thead{\normalsize 1.67 \\  (69.04)} & \thead{\normalsize 1.72 \\  (69.01)}  & \thead{\normalsize 2.77 \\  (69.42)}  \\
\midrule
\thead{\normalsize ResNet50 \\ \small (25.40)} & \normalsize 74.09 & \thead{\normalsize 102.23\\ MB} & \thead{\normalsize 5.45 \\  (73.73)} & \thead{\normalsize \textbf{5.14} \\  (73.65)}  & \thead{\normalsize 5.61\\ (73.73)} & \thead{\normalsize 6.68\\ (73.98)}   \\
\midrule
\thead{\normalsize MobileNet\\ \normalsize-v1 \\ \small (50.73)} &\normalsize 66.18 & \thead{\normalsize 17.02\\ MB}  & \thead{\normalsize 13.29  \\ (66.01)}  & \thead{\normalsize 12.89\\ (66.02)}  &  \thead{\normalsize \textbf{11.16} \\ (65.63)} &  \thead{\normalsize 14.78\\ (65.71)}  \\
\midrule
\thead{\normalsize Small-\\ \normalsize VGG16 \\ \small (7.57)} &\normalsize 91.35 & \thead{\normalsize 60.01 \\ MB} & \thead{\normalsize \textbf{1.90} \\  (91.03)} & \thead{\normalsize 1.95 \\  (91.06)} & \thead{\normalsize 2.08 \\  (91.10) } & \thead{\normalsize 2.84\\  (91.20)} \\
\midrule
\thead{\normalsize LeNet5 \\ \small(1.90)} & \normalsize 99.22 &  \thead{\normalsize 1722\\ KB}  & \thead{\normalsize 0.88 \\ (99.14)} & \thead{\normalsize \textbf{0.87} \\ (99.02)} & \thead{\normalsize 1.09 \\ (99.25)} & \thead{\normalsize 3.01 \\ (99.22)}  \\
\midrule
\thead{\normalsize LeNet-300\\ \normalsize -100 \\ \small (9.05)} & \normalsize 98.29 & \thead{\normalsize 1066\\ KB}  & \thead{\normalsize 2.26 \\  (98.00)}  & \thead{\normalsize 2.20 \\  (98.00)} & \thead{\normalsize \textbf{1.69}\\ (97.76)} &  \thead{\normalsize 4.17\\ (98.36)}  \\
\bottomrule
\end{tabular}
}
\label{Tbl: experimental results}
\end{table}

\footnotetext{Although a better compression ratio was attained, we were not able to get an accuracy in the $\pm{}0.5$ percentage point range of the original accuracy. Therefore, this result shall not be considered as the best result.}

Here we benchmark the maximum compression gains attained by applying DeepCABAC. In order to assess its universality, we applied it to a wide set of pretrained network architectures, trained on different data sets. Concretely, we used the VGG16, ResNet50 and MobileNet-v1 architectures, trained on the ImageNet dataset, a smaller version of the VGG16 architecture trained on the CIFAR10 dataset\footnote{\url{http://torch.ch/blog/2015/07/30/cifar.html}}, which we denote as \textit{Small-VGG16}, and the LeNet-300-100 and LeNet5 trained on MNIST.

In addition, we also applied DeepCABAC to pre-sparsified versions of these networks. For that, we employed the variational sparsification algorithm \cite{VD_sparsifies} to all networks, except for the VGG16 and ResNet50 due to the high computational complexity demanded by the method. The advantage of employing \cite{VD_sparsifies} is that we obtain the variances of each weight parameters as a byproduct of the methods output, thus being able to directly apply DC-v1 after the sparsification process finished. In the cases of the VGG16 and ResNet50 networks, we firstly applied the iterative sparsification algorithm \cite{Learning_Weight_and_Connections} on them and subsequently estimated their FIM-diagonals by minimizing the same variational objective proposed in  \cite{VD_sparsifies} (see appendix for a more in comprehensive explanation).

We compare the two versions of DeepCABAC, DC-v1 \& DC-v2, against two previously proposed quantization schemes. Namely, similarly to \cite{towards_limit_quantization_DNN, Universal_dnn_compression, deep_compression}, we applied the nearest-neighbor quantization scheme on to the networks. In addition, we also applied the weighted Lloyd algorithm as proposed by \cite{weighted_entropy_quantization_DNN, towards_limit_quantization_DNN, Universal_dnn_compression}. As possible lossless compression candidates, we considered the scalar Huffman code, the code proposed by \cite{deep_compression} which we denote \textit{CSR-Huffman}, and the bzip2 \cite{bzip2} algorithm. See appendix for a more detailed explanation of the respective implementations.

Table \ref{Tbl: experimental results} shows the results. As one can see, DeepCABAC is able to attain higher compression gains on most networks as compared to the previously proposed coders. It is able to compress the pretrained by \textbf{x18.9} and the sparsified models by \textbf{x50.6} on average. In contrast, the Lloyd algorithm compresses the models by x13.6 and x47.3 on average, whereas uniform quantization only achieves x5.7 and x25.0 compression gains.

%As a side note we want to stress that we restricted the total amount of evaluations performed by DC-v1 \& DC-v2 to be less than 700 on the respective validation sets. That is, we selected the set of hyperparameters of the quantizer such that maximally 700 evaluations were performed in total on the validation set. This roughly corresponds to restricting the computational complexity of DeepCABAC to be lower than performing 10 training epochs for the ImageNet case.

\subsection{Assignment vs. quantization points}
\begin{table}[!t]
%% increase table row spacing, adjust to taste
\centering
\caption{Average bit-sizes per parameter for the Small-VGG16 network after applying different quantizers. \textit{DC-v1} \& \textit{DC-v2} denote the two versions of DeepCABAC, whereas \textit{Lloyd} denotes the weighted Lloyd algorithm and \textit{uniform} corresponds to the nearest-neighbor quantization. We chose the networks that resided within the $\pm{}0.1$ percentage point range from the accuracy attained after applying a uniform quantizer. In the case of the Lloyd and uniform quantizers, the sizes of the quantized networks were measured with regards to the entropy of their empirical probability mass distribution. In contrast, we measured the explicit average bit-size per parameter in DC-v1 and DC-v2.}
\vspace{0.1in}

\begin{tabular}{M{1.2cm}|M{1.0cm}M{1.0cm}M{1.0cm}M{1.0cm}N}
\toprule
 step-sizes (top1 acc.) & DC-v1 & DC-v2  & Lloyd  & Uniform  \\
\midrule
  \multicolumn{5}{c}{NON-SPARSE} \\

0.032 (90.35) & \textbf{1.48} &  \textbf{1.48} & 1.79 & 1.60  \\

0.016 (91.13)  & 2.21& \textbf{2.20} & 2.29 & 2.40 \\

0.001 (91.55)& 4.27 & 4.80 & \textbf{2.34} & 5.61 \\
\midrule
  \multicolumn{5}{c}{SPARSE (7.57\%)} \\

0.032 (90.22)  & \textbf{0.47} & \textbf{0.47}  & 0.52 & 0.48 \\

0.016 (91.06) & 0.59 & \textbf{0.58} & 0.62 & 0.60 \\

0.001 (91.17) & 0.91 & 1.00 & \textbf{0.74} & 1.00 \\
\bottomrule
\end{tabular}
\label{Tbl: assignment results}
\end{table}

To recall, lossy quantization involves two types of mappings, the quantization map $Q$ where input values are assigned to integers, and the reconstruction map $Q^{-1}$ which assigns a quantization point to each integer. Hence, the following experiment aims to assess the effectiveness of these components individually.

For that, we selected a predefined set of step-sizes and subsequently quantized the parameters according to different quantization schemes. In this way we can attain insights into the compression gains attained only by the influence of the quantization map.

Table \ref{Tbl: assignment results} shows the average bit-sizes per parameter attained by applying different quantizers with the three-given step-sizes to the Small-VGG16 model. In order to decouple the lossless part from the quantization, the bit-sizes are calculated with regards to the entropy of the empirical probability mass distribution (EPMD) of the quantized models in the case of the Lloyd and uniform algorithms, since it marks the theoretical minimum for lossless codes that do not take correlations between the parameters into account. In contrast, since DeepCABAC's quantizer is optimized explicitly under CABACs lossless coder, we calculated the average bit-size with regards to the total bit-size of the model as outputted by CABAC. We also want to stress that we chose the networks that resulted in having equal accuracies, thus, within the $\pm{}0.1$ percentage point range from the accuracy attained after applying a uniform quantizer.

We attain many insights from table \ref{Tbl: assignment results}. Firstly, notice how DeepCABAC's performance is very sensitive to the particular choice of the step-size. This is due to fact that, usually, the best compression performances are attained for small compression strengths $\lambda \approx 0$ at high accuracies. Thus, DeepCABAC's quantization map behaves similarly to uniform quantization in those cases, and therefore it becomes sensitive to the particular choice of the step-size. Nevertheless, notice how DeepCABAC still always attains better compression performance than uniform quantization.

Secondly, notice how for small step-sizes DC-v1 outperforms DC-v2 and thus, makes better rate-distortion decisions. We attribute this to the property that DC-v1 takes the ``robustness'' to perturbations of each element during quantization into account. As we have already discussed in sections \ref{subsec: model compression} and \ref{sec: DeepCABAC} (and in more detail in the appendix), the latter can be interpreted as minimizing an approximation to the desired MDL-loss function. However, since it is only an approximation, this only applies for small step-sizes and becomes more inaccurate for larger ones. Indeed, table \ref{Tbl: assignment results} shows how DC-v2 attains similar results as DC-v1 as the step-size is increased, implying that the particular expression of the RD-function becomes more and more irrelevant.

These insights motivated the design of DC-v2 in the first place, since it is able to explore a larger set of step-sizes for the best accuracy vs. bit-size trade-offs. Indeed, as table \ref{Tbl: experimental results} from the previous experiment shows, DC-v2 attains similar or even higher compression gains than DC-v1, in particular in the case of pretrained networks.

\subsection{Lossless coding}
\begin{table}[!t]
%% increase table row spacing, adjust to taste
\centering
\caption{Compression ratios achieved from lossless compressing different quantized versions of the Small-VGG16 network (and its sparse version). The network was quantized in three different manners, one by applying DC-v2, another with the weighted Lloyd algorithm, and finally with the uniform quantization (nearest-neighbor quantization). The top1 accuracy of each quantized model lies within the $\pm{}0.1$ percentage point range from the original accuracy of the model, which is 91.54\% and 91.35\% respectively. Subsequently, each of them was compressed by applying the scalar Huffman code, the CSR-Huffman code  \cite{deep_compression}, the bzip2 coder, and by CABAC. The second last row denotes the entropy of the EPMD.}
\vspace{0.1in}

\begin{tabular}{M{1.2cm}|M{1.0cm}M{1.0cm}M{1.0cm}N}
\toprule
Quantizers$\rightarrow$  & Uniform & Lloyd & DC-v2 \\
Lossless codes $\downarrow$ & & &  \\
\midrule
\multicolumn{4}{c}{NON-SPARSE} \\

scalar-Huffman & 5.18  & 3.19 & 2.33 \\

  bzip2 & 5.22  & 3.22  & 2.42 \\

    CABAC & \textbf{4.77} & \textbf{2.74}  &  \textbf{2.07} \\

     H & 5.09 & 2.91 & 2.20 \\

\midrule
  \multicolumn{4}{c}{SPARSE (7.57\%)} \\

     scalar-Huffman & 1.35 & 1.71 & 1.33  \\

  CSR-Huffman & 0.91& 0.67 & 0.65 \\

  bzip2 & 0.73 & 0.72 & 0.71 \\

    CABAC & \textbf{0.63} & \textbf{0.63} & \textbf{0.61} \\

     H & 0.84 & 0.60 & 0.58 \\

\bottomrule
\end{tabular}
\label{Tbl: lossless results}
\end{table}

In our last experiment we aimed to assess the efficiency of different universal lossless coders. For that, we quantized the Small-VGG network using three different quantizers, and subsequently compressed each of them using different universal lossless coders. More concretely, we quantized the model by applying DC-v2, the weighted Lloyd algorithm and the nearest-neighbor quantizer. We then applied the scalar Huffman code, the CSR-Huffman code \cite{deep_compression}, the bzip2 algorithm, and CABAC. Moreover, we also calculated  the entropy of the quantized networks, as measured with regards to their empirical probability mass distribution (EPMD).

The resulting bit-sizes are in table \ref{Tbl: lossless results}. As one can see, CABAC is able to attain higher compression gains across all quantized versions of the network. Moreover, in some cases it is able to attain \textbf{lower code-lengths than the entropy of the EPMD}. These results are attributed to CABACs inherent capability to take correlations between the network's parameters into account. This  property highlights its superiority as compared to the previously proposed universal lossless coders, e.g., scalar Huffman and CSR-Huffman, since their average code-lengths are  bounded by the entropy and therefore it would be \textbf{impossible for them to attain lower code-lengths than CABAC}.

\section{Conclusion}
In this work we proposed a novel compression algorithm for deep neural networks called \textit{DeepCABAC}, that is based on applying a Context-based Adaptive Binary Arithmetic Coder (CABAC) to the networks parameters, which is the state-of-the-art universal lossless coder employed in the H.264/HEVC and H.265/HEVC video coding standards. DeepCABAC also incorporates a novel quantization scheme that explicitly minimizes the accuracy vs. bit-size trade-off, without relying on expensive retraining or access to large amounts of data. Experiments showed that it can compress pretrained neural networks by x18.9 on average, and their sparsified versions by x50.6, consistently attaining higher compression performance than previously proposed coding techniques with similar characteristics.  Moreover, DeepCABAC is able to capture correlations between the network's parameters, as such being able to compress the networks parameters beyond the entropy limit of codes that only assume a stationary distribution.

As future work we will investigate the impact of compression on the neural network's problem-solving strategies \cite{LapNCOMM19} and apply DeepCABAC in distributed training scenarios, where communication overhead of the networks update parameters is critical for the overall training efficiency.

\bibliography{References}
\bibliographystyle{IEEEtran}

\clearpage
\appendices
\section{Experiment details}
\subsection{Uniform quantization}
Uniform quantization is essentially one step of the weighted Lloyd algorithm with no importance measure, $\lambda{} = 0$, and no cluster center update. One major difference between uniform quantization and the weighted Lloyd algorithm is, that in the weighted Lloyd algorithm, the neural network is quantized as a whole, while in uniform quantization, the neural network is quantized layer-wise (see algorithm \ref{algorithm:uniform-quantization-algorithm}).

For the experiment described in section \ref{subsec: Exp1}, the number of clusters were determined by first starting out with 256 clusters for un-sparsified networks and with 32 clusters for sparsified networks. Then the networks were quantized and evaluated. If the accuracy was not within a range of $\pm{}0.5$ percentage points as compared to the original accuracy, then the number of clusters was doubled until the accuracy was within that range.

\begin{table}
    \centering
    \caption{The number of clusters per layer employed for the experiment described in section \ref{subsec: Exp1}.}
    \begin{tabular}{l | r}
        \toprule
        Model & Clusters \\
        \midrule
        LeNet-300-100 & 256 \\
        LeNet-300-100 Sparse & 32 \\
        \midrule
        LeNet5 & 256 \\
        LeNet5 Sparse & 32 \\
        \midrule
        Small-VGG16 & 256 \\
        Small-VGG16 Sparse & 128 \\
        \midrule
        ResNet50 & 256 \\
        ResNet50 Sparse & 256 \\
        \midrule
        VGG16 & 256 \\
        VGG16 Sparse & 256 \\
        \midrule
        MobileNet v1 & 1,024 \\
        MobileNet v1 Sparse & 1,024 \\
        \bottomrule
    \end{tabular}
\end{table}

The quantized models were then compressed using scalar Huffman coding and bzip2. Additionally, the sparse models were compressed using CSR-Huffman coding. Since additional parameters such as biases were not quantized, their original size was added to the compressed size of all compression methods.

\subsection{Lloyds hyperparameter selection}

To determine the optimal number of clusters and optimal values for $\lambda{}$, in the beginning, the number of clusters were fixed to 256. The ranges for $\lambda{}$ were determined iteratively under the assumption that the accuracy decreases roughly monotonic with an increasing $\lambda{}$. At first, 20 experiments with a $\lambda{}$ value between 0.0 and 1.0 were started and evaluated to establish a rough range of $\lambda{}$ in which the accuracies are within a range of $\pm{}0.5$ percentage points as compared to its original accuracy. In one case (LeNet-300-100 Sparse) all experiments yielded accuracies within that range. In that case another 20 experiments were conducted with $\lambda{}$ values between 1.0 and 2.0. In the case of MobileNet v1 and its sparse counterpart, even a value of 0.0 for $\lambda{}$ produced accuracies below the $\pm{}0.5$ percentage point threshold. In both cases the number of clusters was doubled, then 20 experiments with $\lambda{}$ in the range of 0.0 to 1.0 were conducted. This process was repeated until the accuracies were within the threshold\footnote{Please note that, although the number of clusters was drastically increased for MobileNet v1, we were not able to achieve accuracies within the target threshold.}. Then for all networks two adjacent values of $\lambda{}$ were selected where the accuracy lies within the range and that produced the smallest entropies. Again, 20 experiments with $\lambda{}$ values between these selected $\lambda{}$s were conducted. This process was repeated until there were no longer any gains in the entropies. Typically, only two rounds were enough to find no further improvement.

\begin{table}
    \caption{The number of clusters for the whole network as well as the $\lambda{}$ values used in experiment \ref{subsec: Exp1}}
    \centering{}
    \begin{tabular}{l | r | r}
        \toprule
        Model & \multicolumn{1}{c|}{$\lambda{}$} & Clusters \\
        \midrule
        LeNet-300-100 & 0.0144 & 256 \\
        LeNet-300-100 Sparse & 1.1053 & 256 \\
        \midrule
        LeNet5 & 0.1222 & 256 \\
        LeNet5 Sparse & 0.4 & 256 \\
        \midrule
        Small-VGG16 & 0.2105 & 256 \\
        Small-VGG16 Sparse & 0.2368 & 256 \\
        \midrule
        ResNet50 & 0.05 & 256 \\
        ResNet50 Sparse & 0.0105 & 256 \\
        \midrule
        VGG16 & 0.0063 & 256 \\
        VGG16 Sparse & 0.0474 & 256 \\
        \midrule
        MobileNet v1 & 0.9474 & 10,240 \\
        MobileNet v1 Sparse & 0.9474 & 3,072 \\
        \bottomrule
    \end{tabular}
\end{table}

\subsection{DeepCABAC's hyperparameter selection}
In all experiments we set the \textit{AbsGr(n)}-Flag to 10. 

\subsection{DC-v1}
For the experiment in section \ref{subsec: Exp1}, we searched through the following set of hyperparameters:
\begin{align*}
S = 	& [0.0, 8.0, 16.0, 32.0, 64.0, 96.0, 128.0, \\
 	& 160.0, 172.0, 192.0, 256.0] \\
\lambda = & 0.0001\cdot 2^{(\log_2{(10^2)} \frac{i}{100})}, \quad \forall i\in \{0,...,99\}
\end{align*}

\subsection{DC-v2}
For the experiment in section \ref{subsec: Exp1}, we searched through the following set of hyperparameters:
\begin{align*}
\lambda = & \frac{0.02}{20}\cdot i +0.01, \quad \forall  i\in \{0,...,20\} \\ 
\Delta_1 = & 0.001\cdot 2^{(\log_2{\left(\frac{0.15}{0.001}\right)} \frac{i}{70})}, \quad \forall i\in \{0,...,70\} \\
\Delta_2 = & 0.064\cdot 2^{(\log_2{\left(\frac{0.128}{0.064}\right)} \frac{i}{30})}, \quad \forall i\in \{0,...,30\}
\end{align*}

\section{Approximations to the Fisher Information Matrix}
\cite{VD_sparsifies} proposed a sparsification algorithm for neural networks, that is based on the minimization of a variational objective. Concretely, they assume the improper log-scale uniform prior $P(w)$ and assume a fully factorized  gaussian posterior over the weight parameters $P(w|\mu, \sigma)$, and minimize the corresponding variational upper bound
\begin{equation}
(\mu, \sigma)^* = \min_{(\mu, \sigma)} \mathbb{E}_{P(w|\mu, \sigma)}[\mathcal{L}(y, y')] + \beta D_{KL}(P(w|\mu, \sigma)||P(w))
\label{Eq: VD sparsifies variational upper bound}
\end{equation}
with $y' \sim P(y'|x, w)$ being the output samples of the neural network, $(x,y)$ the data samples, and $(\mu, \sigma)$ the mean and standard deviations of all the networks parameters, and $\beta \in \mathbb{R}$ the Lagrange-multiplier. As the KL-Divergence cannot be calculated analytically, they proposed to approximate it by
\begin{align}
 & D_{KL}(P(w|\mu, \sigma)||P(w)) \nonumber \\
 \approx &  \sum_i k_1 \text{sgm}(k_2 + k_3\log\alpha_i) -\frac{1}{2} \log{\left(\frac{\alpha_i +1}{\alpha_i}\right)}
\label{Eq: KL-approx}
\end{align}
with $ \text{sgm}(\cdot)$ being the sigmoid function, $\alpha_i = \sigma^2_i/\mu^2_i$ the inverse of the signal-to-noise ratio of the parameters, $k_1 = 0.63576$, $k_2 = 1.87320$ and $k_3 = 1.48695$. Then, they minimize \eqref{Eq: VD sparsifies variational upper bound} by applying scalable sampling techniques proposed by \cite{variational_dropout}.

\subsection{Connection between pruning and quantization}
As a result of minimizing  \eqref{Eq: VD sparsifies variational upper bound} we get a mean and standard deviation for each parameter of the network. In our work, we interpreted the former as their (new) value and the latter as a measure of their ``robustness'' against perturbations. Indeed, the authors suggested to prune away (set to 0) parameters with a signal-to-noise-ratio under a given threshold. Concretely, they suggested the following pruning scheme
\[w_i \rightarrow 0, \quad \forall i: \alpha^{-1} < e^{-3}\]
where $w_i\equiv \mu_i$ represents now the mean value and thus $\alpha^{-1} = w_i^2/\sigma_i^2$.

We can see that the scalar rate-distortion objective \eqref{Eq: scalar diag fisher rate-distortion function}
\begin{equation*}
(Q, Q^{-1})^*=  \min_{(Q, Q^{-1})}\; F_i(q_i-w_i)^2 + \lambda L_Q(b)
\end{equation*}
is a generalization of the above sparsification scheme. Namely, if we assume that the set of quantization points entails the same elements as the input set $\mathbb{W} = \mathbb{Q}$ (thus, $Q^{-1}\equiv$ identity map), and consider a decoder that assumes a spike-and-slab distribution over the quantization points, then the above objective can be solved by applying the Lloyd algorithm. After convergence, it results in the following solution
\begin{equation*}
Q^*(w_i)= 0\quad \text{if} \quad  F_iw_i^2  < \lambda (b+\log_2p_0),
\end{equation*}
with $p_0$ being the empirical probability distribution of the 0 value and $b$ the bit-precision for representing the non-zero values. Hence, if we now choose $F_i = 1/\sigma_i^2$ and the adequate $\lambda$, we get the suggested criteria as a special case solution. This insight motivated our choice of FIM-diagonals in our experimental section.

\subsection{Connection between variances, Hessian, and FIM-diagonals}
Firstly, as thoroughly discussed in \cite{Natural_gradient_method} and mentioned in \cite{CLP_achile}, it is important to recall that the FIM is a semi-positive approximation of the Hessian of the loss function.

Hence, similar to \cite{CLP_achile}, we can derive a more rigorous connection between the estimated variances from minimizing \eqref{Eq: VD sparsifies variational upper bound}, the FIM-diagonals and the Hessian. Namely, assuming that the variational loss function can be approximated by its second order expansion around the weight configuration $w$, we get the following expression
\[\mathbb{E}_{P(w|\mu, \sigma)}[\mathcal{L}(y, y')] \approx \mathcal{L}(w) +  \frac{1}{2}\text{tr}[\sigma H(w)]\]
with  $\mathcal{L}(w)$ being the loss value at $w$ and $\text{tr}[\cdot]$ the trace. Hence, if we substitute this expression into \eqref{Eq: VD sparsifies variational upper bound} and take the derivative with respect to $\sigma_i$ we attain
\begin{equation*}
H_i = \frac{\beta}{\sigma_i^2}K(\alpha^{-1})
\end{equation*}
with $K(\alpha^{-1}) \in (0,1)$ being (approximately) a monotonically increasing function of the signal-to-noise ratio of the parameter. Hence, there is a direct connection between the variances, signal-to-noise ratio, and hessian of the loss function and, consequently, with the FIM-diagonals.

\subsection{Hessian-weighted vs. variance-weighted quantization}
\begin{figure}[t]
    \centering
    \includegraphics[width=0.75\columnwidth,clip,keepaspectratio]{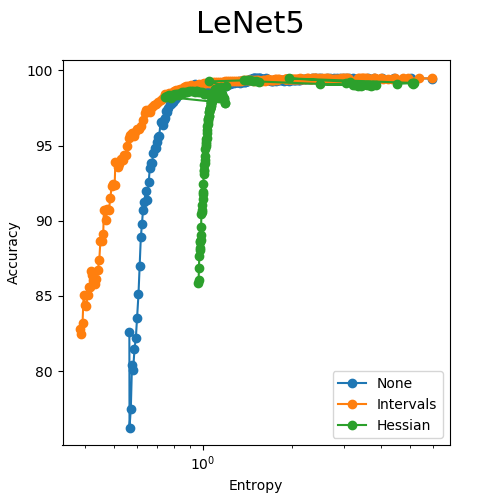}
    \caption{Rate-accuracy curves for a pretrained LeNet5 model after applying the weighted Lloyd algorithm with different importance measures.}
    \label{Fig: Lenet5 RD}
\end{figure}

\cite{towards_limit_quantization_DNN} suggested a Hessian-weighted Lloyd algorithm for quantizing the neural networks parameters, where the diagonals of the empirical Hessian are taken as weights in the algorithm. As we have already discussed above, these coefficients are closely connected to the FIM-diagonals, and are thus also theoretically well motivated. However, we experienced the algorithm to be less stable in practice when we used the Hessian-diagonals instead of the variances. Figure \ref{Fig: Lenet5 RD} shows the rate-accuracy curves when we quantized the LeNet5 model with both alternatives. As we can see, the curves of the variances are more stable, even achieving better compression results than the Hessian-weighted variant.

\section{Pseudocodes}

\begin{algorithm}
  \caption{Encoding a message using a Huffman code.}
  \label{algorithm:huffman-encoding}

  \begin{algorithmic}[1]
      \State{\textbf{Input:} A message $M \coloneqq{} \left ( s_{1}, s_{2}, \dots{}, s_{n} \right )$ of length $n$, over alphabet $\Sigma{}$ and a Huffman code $H: \Sigma{} \rightarrow{} \left \{ 0, 1 \right \}^{+}$}
      \State{\textbf{Output:} A sequence of bits $E \in{} \left \{ 0, 1 \right \}^{+}$ containing the encoded symbol sequence}

      \State{Let $E = \varepsilon{}$}
      \ForAll{$s \in{} M$}
          \State{$E \leftarrow{} E \circ{} H_{s}$}
      \EndFor{}

      \State{\Return{$E$}}
  \end{algorithmic}
\end{algorithm}

\begin{algorithm}
  \caption{Decoding a message using a Huffman code.}
  \label{algorithm:huffman-decoding}

  \begin{algorithmic}[1]
      \State{\textbf{Input:} A sequence of bits $E \in{} \left \{ 0, 1 \right \}^{+}$ to decode and a Huffman code $H: \Sigma{} \rightarrow{} \left \{ 0, 1 \right \}^{+}$}
      \State{\textbf{Output:} A message $M \coloneqq{} \left ( s_{1}, s_{2}, \dots{}, s_{n} \right )$ of length $n$, over alphabet $\Sigma{}$ representing the decoded message}

      \State{Let $M = \varepsilon{}$}
      \While{$\left | E \right | > 0$}
          \ForAll{$\sigma{} \in{} \Sigma{}$}
              \If{$\exists{} x \in \Sigma{}^{*}: E = H_{\sigma{}} \circ{} x$}
                  \State{$M \leftarrow{} M \circ{} \sigma{}$}
                  \State{$E \leftarrow{} (E_{|E_{\sigma{}}|}, \dots{}, E_{|E|-|E_{\sigma{}}|})$}
              \EndIf{}
          \EndFor{}
      \EndWhile{}

      \State{\Return{$M$}}
  \end{algorithmic}
\end{algorithm}

\begin{algorithm}
  \caption{Generating a Huffman code for a message}
  \label{algorithm:generating-huffman-code}

  \begin{algorithmic}[1]
      \State{\textbf{Input:} A message $M \coloneqq{} \left ( s_{1}, s_{2}, \dots{}, s_{n} \right )$ of length $n$, over alphabet $\Sigma{}$}
      \State{\textbf{Output:} A Huffman code, which maps symbols to code words: $H: \Sigma{} \rightarrow{} \left \{ 0, 1 \right \}^{+}$}

      \ForAll{$\sigma{} \in{} \Sigma{}$} \Comment{Calculate symbol frequency}
          \State{$F_{\sigma{}} \leftarrow{} \frac{\left | \left ( s \in{} M \; | \; s = \sigma{} \right ) \right |}{\left | M \right |}$}
      \EndFor{}

      \State{Let $Q$ be a priority queue} \Comment{Dequeuing from $Q$ yields smallest element}
      \ForAll{$s \in{} M$}
          \If{$F_{s} > 0$}
              \State{$\Call{Enqueue}{\left (s, F_{s} \right), Q}$}
          \EndIf{}
      \EndFor{}

      \While{$\left | Q \right | > 1$} \Comment{Build Huffman tree}
          \State{$(n_{l}, F_{l}) \leftarrow{} \Call{Dequeue}{Q}$}
          \State{$(n_{r}, F_{r}) \leftarrow{} \Call{Dequeue}{Q}$}
          \State{$\Call{Enqueue}{\left ((n_{l}, n_{r}), F_{l} + F_{r} \right), Q}$}
      \EndWhile{}

      \State{$(n, F_{n}) \leftarrow{} \Call{Dequeue}{Q}$}
      \State{\Return{\Call{AssignCodeWords}{$n$, $\varepsilon{}$}}}

      \Procedure{AssignCodeWords}{$n$, $c$}
          \If{$n$ \textbf{is} leaf node} \Comment{Assign code word to leaf nodes}
              \State{$H_{n} \leftarrow{} c$}
          \Else{} \Comment{Traverse tree recursively}
              \State{$H \leftarrow{} H \cup{} \Call{AssignCodeWords}{n_{l}, c \circ{} 0}$}
              \State{$H \leftarrow{} H \cup{} \Call{AssignCodeWords}{n_{r}, c \circ{} 1}$}
          \EndIf{}

          \State{\Return{H}}
      \EndProcedure{}
  \end{algorithmic}
\end{algorithm}

\begin{algorithm}
  \caption{Weighted Lloyd's algorithm}
  \label{algorithm:weighted-lloyds-algorithm}

  \begin{algorithmic}[1]
      \State{\textbf{Input:} A set of clusters $C = \left \{ c_{1}, c_{2}, \dots{}, c_{k} \right \}$,
          a set of neural network parameters $W = \left \{ w_{1}, w_{2}, \dots{}, w_{n} \right \}$,
          an importance measure for each neural network parameter
          $F = \left \{ f_{1}, f_{2}, \dots{}, f_{n} \right \}$, and a Lagrangian multiplier
          $\lambda{}$
      }
      \State{\textbf{Output:} A set of cluster assignments
          $\mathcal{C}_{j} = \left \{ w_{1}, w_{2}, \dots{}, w_{i} \right \} \forall{} j \in{} \left \{ 1, \dots{}, k \right \}$
          that minimize the Lagrangian loss function $J_{\lambda{}}$
      }

      \For{$j = 1 \rightarrow{} k$} \Comment{Initialize cluster probabilities}
          \State{$P_{j} \leftarrow{} \frac{1}{k}$}
      \EndFor{}

      \Repeat{}
          \For{$j = 1 \rightarrow{} k$} \Comment{Reset the clusters}
              \State{$\mathcal{C}_{j} \leftarrow{} \varnothing{}$}
          \EndFor{}

          \For{$i = 1 \rightarrow{} n$} \Comment{Assignment step}
              \State{$j = \underset{j}{\mathrm{argmin}} \left ( F_{i} (W_{i} - C_{j})^{2} - \lambda{} log_{2}(P_{j}) \right )$}
              \State{$\mathcal{C}_{j} \leftarrow{} \mathcal{C}_{j} \cup{} \left \{ W_{i} \right \}$}
          \EndFor{}

          \For{$j = 1 \rightarrow{} k$} \Comment{Update step}
              \State{$C_{j} \leftarrow{} \frac{\sum_{w_{i} \in{} \mathcal{C}_{j}} F_{i}w_{i}}{\sum_{w_{i} \in{} \mathcal{C}_{j}} F_{i}}$}
              \State{$P_{j} \leftarrow{} \frac{\left | \mathcal{C}_{j} \right |}{n}$}
          \EndFor{}

          \State{$j = \underset{j}{\mathrm{argmin}} \left | C_{j} \right |$} \Comment{Enforce 0-cluster}
          \State{$C_{j} \leftarrow{} 0$}

          \State{$J_{\lambda{}} = \sum_{j = 1}^{k} \sum_{w_{i} \in{} \mathcal{C}_{j}} F_{i} (w_{i} - C_{j})^{2} - \lambda{} log_{2}(P_{j})$}
          \Comment{Calculate loss}

      \Until{convergence, e.g. $J_{\lambda{}}$ decreases below some threshold}

      \State{\Return{$\mathcal{C}$}}
  \end{algorithmic}
\end{algorithm}

\begin{algorithm}
  \caption{Uniform quantization algorithm}
  \label{algorithm:uniform-quantization-algorithm}

  \begin{algorithmic}[1]
      \State{\textbf{Input:} A set of $n$ neural network layers
          $W_{l} = \left \{ w_{1}, w_{2}, \dots{}, w_{n_{l}} \right \} \forall{} l \in{} \left \{ 1, \dots{}, n \right \}$
          and a set of cluster centers for each layer of the neural network:
          $C_{l} = \left \{ c_{1}, c_{2}, \dots{}, c_{k_{l}} \right \} \forall{} l \in{} \left \{ 1, \dots{}, n \right \}$
      }
      \State{\textbf{Output:} A set of cluster assignments
          $\mathcal{C}_{l} = \left \{ w_{1}, w_{2}, \dots{}, w_{i_{l}} \right \} \forall{} l \in{} \left \{ 1, \dots{}, n \right \}$
      }

      \For{$l = 1 \rightarrow{} n$}
          \For{$i = 1 \rightarrow{} n_{l}$}
              \State{$j = \underset{j}{\mathrm{argmin}} \left (W_{li} - C_{lj} \right )^{2}$}
              \State{$\mathcal{C}_{lj} \leftarrow{} \mathcal{C}_{lj} \cup{} \left \{ W_{li} \right \}$}
          \EndFor{}
      \EndFor{}

      \State{\Return{$\mathcal{C}$}}
  \end{algorithmic}
\end{algorithm}

\end{document}